\documentclass[conference,final,]{IEEEtran}

\usepackage{graphicx}
\makeatletter
\def\maxwidth{\ifdim\Gin@nat@width>\linewidth\linewidth
\else\Gin@nat@width\fi}
\makeatother
\let\Oldincludegraphics\includegraphics
\renewcommand{\includegraphics}[1]{\Oldincludegraphics[width=\maxwidth]{#1}}

\usepackage[unicode=true]{hyperref}

\hypersetup{
            pdftitle={The Spatially-Conscious Machine Learning Model},
            pdfkeywords={Real estate, Artificial neural networks, Machine learning, Recommender
systems, Supervised learning, Predictive modeling},
            pdfborder={0 0 0},
            breaklinks=true}
\usepackage{placeins}
\usepackage{fancyhdr}
\usepackage{setspace}
\usepackage{chngcntr}
\usepackage{microtype}
\usepackage{ragged2e}
\usepackage{amsmath}
\usepackage{indentfirst}
\usepackage{longtable}
\usepackage{algorithm}
\usepackage{algpseudocode}
\usepackage{float}
\usepackage{enumitem}
\counterwithin{figure}{section}
\counterwithin{table}{section}
\usepackage{amsmath}
\usepackage{graphicx}

\providecommand{\tightlist}{%
  \setlength{\itemsep}{0pt}\setlength{\parskip}{0pt}}

\providecommand{\keywords}[1]
{
  \small	
  \textbf{\textit{Keywords---}} #1
}

\begin{document}
\title{The Spatially-Conscious Machine Learning Model}


\author{


\IEEEauthorblockN{
Timothy J. Kiely 
}
\IEEEauthorblockA{Northwestern University\\
School of Professional Studies\\
Chicago, IL
\\timothy.j.kiely@gmail.com
}
\and
\IEEEauthorblockN{
Dr.~Nathaniel D. Bastian 
}
\IEEEauthorblockA{U.S. Military Academy\\
Army Cyber Institute\\
West Point, NY
\\nathaniel.bastian@westpoint.edu
}


}

\maketitle

\begin{abstract}
Successfully predicting gentrification could have many social and
commercial applications; however, real estate sales are difficult to
predict because they belong to a chaotic system comprised of intrinsic
and extrinsic characteristics, perceived value, and market speculation.
Using New York City real estate as our subject, we combine modern
techniques of data science and machine learning with traditional spatial
analysis to create robust real estate prediction models for both
classification and regression tasks. We compare several cutting edge
machine learning algorithms across spatial, semi-spatial and non-spatial
feature engineering techniques, and we empirically show that
spatially-conscious machine learning models outperform non-spatial
models when married with advanced prediction techniques such as
feed-forward artificial neural networks and gradient boosting machine
models.
\end{abstract}



\maketitle


%
\IEEEpeerreviewmaketitle

\hypertarget{introduction}{%
\section{Introduction}\label{introduction}}

Things near each other tend to be like each other. This concept is a
well-known problem in traditional spatial analysis and is typically
referred to as \emph{spatial autocorrelation}. In statistics, this is
said to ``reduce the amount of information'' pertaining to spatially
proximate observations as they can, in part, be used to predict each
other (DiMaggio, 2012). But can spatial features be used in a machine
learning context to make better predictions? This work demonstrates that
the addition of ``spatial lag'' features to machine learning models
significantly increases accuracy when predicting real estate sales and
sale prices.

\hypertarget{application-combating-income-inequality-by-predicting-gentrification}{%
\subsection{Application: Combating Income Inequality by Predicting
Gentrification}\label{application-combating-income-inequality-by-predicting-gentrification}}

Researchers at the Urban Institute (Greene, Pendall, Scott, \& Lei,
2016) recently identified \emph{economic exclusion} as a powerful
contributor to income inequality in the United States. Economic
exclusion can be defined as follows: vulnerable
populations--disproportionately communities of color, immigrants,
refugees, and women--who are geographically segregated from economic
prosperity enter a continuous cycle of diminished access to good jobs,
good schools, health care facilities, public spaces, and other economic
and social resources. Diminished access leads to more poverty, which
leads to more exclusion. This self-reinforcing cycle of poverty and
exclusion gradually exacerbates income inequality over the course of
years and generations.

Economic exclusion typically unfolds as a byproduct of gentrification.
When an area experiences economic growth, increased housing demands and
subsequent affordability pressures can lead to voluntary or involuntary
relocation of low-income families and small businesses. To prevent
economic exclusion, it is necessary to prevent this negative consequence
of gentrification, known as \emph{displacement}, (Clay, 1979). What can
be done to intervene?

Efforts by government agencies and nonprofits to intervene typically
occur once displacement is already underway, and after-the-fact
interventions can be costly and ineffective. Several preemptive actions
exist which can be deployed to stem divestment and ensure that existing
residents benefit from local prosperity. Potential interventions include
job training, apprenticeships, subsidies, zoning laws, charitable aid,
matched savings programs, financial literacy coaching, homeowner
assistance, housing vouchers, and more (Greene et al., 2016). Yet not
unlike medical treatment, early detection is the key to success.
Reliably predicting gentrification would be a valuable tool for
preventing displacement at an early stage; however, such a task has
proven difficult historically.

One response to this problem has been the application of predictive
modeling to forecast likely trends in gentrification. The Urban
Institute published a series of essays in 2016 outlining the few ways
city governments employ ``Big data and crowdsourced data'' to identify
vulnerable individuals and connect them with the proper services and
resources, noting that ``much more could be done'' (Greene et al.,
2016).

To date, many government agencies have demonstrated the benefits of
applied predictive modeling, ranging from prescription drug abuse
prevention to homelessness intervention to recidivism reduction (Ritter,
2013). However, few if any examples exist of large-scale, systematic
applications of data analysis to aid vulnerable populations experiencing
displacement. This work belongs to an emerging trend known as the
``science of cities'' which aims to use large data sets and advanced
simulation and modeling techniques to understand and improve urban
patterns and how cities function (Batty, 2013).

Below we describe techniques that can dramatically boost the accuracy of
existing gentrification prediction models. We use real estate
transactions in New York City, both their occurrence (probability of
sale) and their dollar amount (sale price per square foot) as a proxy
for gentrification. The technique marries the use of machine learning
predictive modeling with spatial lag features typically seen in
geographically-weighted regressions (GWR). We employ a two-step modeling
process in which we 1) determine the optimal building types and
geographies suited to our feature engineering assumptions and 2) perform
a comparative analysis across several state-of-the-art algorithms
(generalized linear model, Random Forest, gradient boosting machine, and
artificial neural network). We conclude that spatially-conscious machine
learning models consistently outperform traditional real estate
valuation and predictive modeling techniques.

\hypertarget{literature-review}{%
\section{Literature Review}\label{literature-review}}

This literature review discusses the academic study of economic
displacement, primarily as it relates to gentrification. We also examine
\emph{mass appraisal techniques}, which are automated analytical
techniques used for valuing large numbers of real estate properties.
Finally, we examine recent applications of machine learning as it
relates to predicting gentrification.

\hypertarget{what-is-economic-displacement}{%
\subsection{What is Economic
Displacement?}\label{what-is-economic-displacement}}

Economic displacement has been intertwined with the study of
gentrification since shortly after the latter became academically
relevant in the 1960s. The term gentrification was first introduced in
1964 to describe the \emph{gentry} in low-income neighborhoods in London
(Glass, 1964). Initially, academics described gentrification in
predominantly favorable terms as a ``tool of revitalization'' for
declining neighborhoods (Zuk et al., 2015). However, by 1979 the
negative consequences of gentrification became better understood,
especially with regards to economic exclusion (Clay, 1979). Today, the
term has a more neutral connotation, describing the placement and
distribution of populations (Zuk et al., 2015). Specific to cities,
recent literature defines gentrification as the process of transforming
vacant and working-class areas into middle-class, residential or
commercial areas (Chapple \& Zuk, 2016; Lees, Slater, \& Wyly, 2013).

Studies of gentrification and displacement generally take two approaches
in the literature: supply-side and demand-side (Zuk et al., 2015).
Supply-side arguments for gentrification tend to focus on investments
and policies and are much more often the subject of academic literature
on economic displacement. This kind of research may be more common
because it has the advantage of being more directly linked to
influencing public policy. According to Dreier, Mollenkopf, \& Swanstrom
(2004), public policies that can increase economic displacement have
been, among others, automobile-oriented transportation infrastructure
spending and mortgage interest tax deductions for homeowners. Others who
have argued for supply-side gentrification include Smith (1979), who
stated that the return of capital from the suburbs to the city, or the
``political economy of capital flows into urban areas'' are what
primarily drive both the positive and negative consequences of urban
gentrification.

More recently, researchers have explored economic displacement as a
contributor to income inequality (Reardon \& Bischoff, 2011; Watson,
2009). Wealthy households tend to influence local political processes to
reinforce exclusionary practices. The exercising of political influence
by prosperous residents results in a feedback loop producing downward
economic pressure on households who lack such resources and influence.
Gentrification prediction tools could be used to help break such
feedback loops through early identification and intervention.

Many studies conclude that gentrification in most forms leads to
exclusionary economic displacement; however, Zuk et al. (2015)
characterizes the results of many recent studies as ``mixed, due in part
to methodological shortcomings.'' This work attempts to further the
understanding of gentrification prediction by demonstrating a technique
to better predict real estate sales in New York City.

\hypertarget{a-review-of-mass-appraisal-techniques}{%
\subsection{A Review of Mass Appraisal
Techniques}\label{a-review-of-mass-appraisal-techniques}}

Much research on predicting real estate prices has been in service of
creating mass appraisal models. Local governments most commonly use mass
appraisal models to assign taxable values to properties. Mass appraisal
models share many characteristics with predictive machine learning
models in that they are data-driven, standardized methods that employ
statistical testing (Eckert, 1990). A variation on mass appraisal models
are the \emph{automated valuation models} (AVM). Both mass appraisal
models and AVMs seek to estimate the market value of a single property
or several properties through data analysis and statistical modeling
(d'Amato \& Kauko, 2017).

Scientific mass appraisal models date back to 1936 with the reappraisal
of St.~Paul, Minnesota (Joseph, n.d.). Since that time, and accelerating
with the advent of computers, much statistical research has been done
relating property values and rent prices to various characteristics of
those properties, including their surrounding area. Multiple regression
analysis (MRA) has been the most common set of statistical tools used in
mass appraisal, including maximum likelihood, weighted least squares,
and the most popular, ordinary least squares, or OLS (d'Amato \& Kauko,
2017). MRA techniques, in particular, are susceptible to spatial
autocorrelation among residuals. Another group of models that seek to
correct for spatial dependence are known as spatial auto-regressive
models (SAR), chief among them the spatial lag model, which aggregates
weighted summaries of nearby properties to create independent regression
variables (d'Amato \& Kauko, 2017).

So-called \emph{hedonic regression models} seek to decompose the price
of a good based on the intrinsic and extrinsic components. Koschinsky,
Lozano-Gracia, \& Piras (2012) is a recent and thorough discussion of
parametric hedonic regression techniques. Koschinsky derives some of the
variables included in his models from nearby properties, similar to the
techniques used in this work, and these spatial variables were found to
be predictive. The basic real estate hedonic model describes the price
of a given property as:

\[
P_i = P(q_i,S_i, N_i, L_i)
\]

\noindent where \(P_i\) represents the price of house \(i\), \(q_i\)
represents specific environmental factors, \(S_i\) are structural
characteristics, \(N_i\) are neighborhood characteristics, and \(L_i\)
are locational characteristics (Koschinsky et al., 2012 pg. 322).
Specifically, the model calculates spatial lags on properties of
interest using neighboring properties within 1,000 feet of a sale. The
derived variables include characteristics like average age, the number
of poor condition homes, percent of homes with electric heating,
construction grades, and more. Koschinsky found that in all cases homes
near each other were typically similar to each other and priced
accordingly, concluding that locational characteristics should be valued
at least as much ``if not more'' than intrinsic structural
characteristics (Koschinsky et al., 2012).

As recently as 2015, much research has dealt with mitigating the
drawbacks of MRA. Fotheringham, Crespo, \& Yao (2015) explored the
combination of geographically weighted regression (GWR) with time-series
forecasting to predict home prices over time. GWR is a variation on OLS
that assigns weights to observations based on a distance metric.
Fotheringham et al. (2015) successfully used cross-validation to
implement adaptive bandwidths in GWR, i.e., for each observation, the
number of neighboring data points included in its spatial radius were
varied to optimize performance.

\hypertarget{predicting-gentrification-using-machine-learning}{%
\subsection{Predicting Gentrification Using Machine
Learning}\label{predicting-gentrification-using-machine-learning}}

Both mass appraisal techniques and AVMs seek to predict real estate
prices using data and statistical methods; however, traditional
techniques typically fall short. These techniques fail partly because
property valuation is inherently a ``chaotic'' process that cannot be
modeled effectively using linear methods (Zuk et al., 2015). The value
of any given property is a complex combination of fungible intrinsic
characteristics, perceived value, and speculation. The value of any
building or plot of land belongs to a rich network where decisions about
and perceptions of neighboring properties influence the final market
value. Guan, Shi, Zurada, \& Levitan (2014) compared traditional MRA
techniques to alternative data mining techniques resulting in mixed
results. However, as Helbich, Jochem, Mücke, \& Höfle (2013) state,
hedonic pricing models can be improved in two primary ways: through
novel estimation techniques, and by ancillary structural, locational,
and neighborhood variables. Recent research generally falls into these
two buckets: better algorithms and better data.

In the better data category, researchers have been striving to introduce
new independent variables to increase the accuracy of predictive models.
Alexander Dietzel, Braun, \& Schäfers (2014) successfully used internet
search query data provided by Google Trends to serve as a sentiment
indicator and improve commercial real estate forecasting models. Pivo \&
Fisher (2011) examined the effects of walkability on property values and
investment returns. Pivo found that on a 100-point scale, a 10-point
increase in walkability increased property investment values by up to
9\% (Pivo \& Fisher, 2011).

Research into better prediction algorithms and employing better data are
not mutually exclusive. For example, Fu et al. (2014) created a
prediction algorithm, called \emph{ClusRanking}, for real estate in
Beijing, China. ClusRanking first estimates neighborhood characteristics
using taxi cab traffic vector data, including relative access to
business areas. Then, the algorithm performs a rank-ordered prediction
of investment returns segmented into five categories. Similar to
Koschinsky et al. (2012), though less formally stated, Fu et al. (2014)
modeled a property's value as a composite of individual, peer and zone
characteristics by including characteristics of the neighborhood, the
values of nearby properties, and the prosperity of the affiliated latent
business area based on taxi cab data (Fu et al., 2014).

Several other recent studies compare various advanced statistical
techniques and algorithms either to other advanced techniques or to
traditional ones. Most studies conclude that the advanced,
non-parametric techniques outperform traditional parametric techniques,
while several conclude that the Random Forest algorithm is particularly
well-suited to predicting real estate values.

Kontrimas \& Verikas (2011) compared the accuracy of linear regression
against the SVM technique and found the latter to outperform.
Schernthanner, Asche, Gonschorek, \& Scheele (2016) compared traditional
linear regression techniques to several techniques such as kriging
(stochastic interpolation) and Random Forest. They concluded that the
more advanced techniques, particularly Random Forest, are sound and more
accurate when compared to traditional statistical methods. Antipov \&
Pokryshevskaya (2012) came to a similar conclusion about the superiority
of Random Forest for real estate valuation after comparing 10
algorithms: multiple regression, CHAID, exhaustive CHAID, CART, 2 types
of k-nearest neighbors, multilayer perceptron artificial neural network,
radial basis functional neural network, boosted trees and finally Random
Forest.

Guan et al. (2014) compared three different approaches to defining
spatial neighbors: a simple radius technique, a k-nearest neighbors
technique using only distance and a k-nearest neighbors technique using
all attributes. Interestingly, the location-only KNN models performed
best, although by a slight margin. Park \& Bae (2015) developed several
housing-price prediction models based on machine learning algorithms
including C4.5, RIPPER, naive Bayesian, and AdaBoost, finding that the
RIPPER algorithm consistently outperformed the other models. Rafiei \&
Adeli (2015) employed a restricted Boltzmann machine (neural network
with back propagation) to predict the sale price of residential condos
in Tehran, Iran, using a non-mating genetic algorithm for dimensionality
reduction with a focus on computational efficiency. The paper concluded
that two primary strategies help in this regard: weighting property
sales by temporal proximity (i.e., sales which happened closer in time
are more alike), and using a learner to accelerate the recognition of
important features.

Finally, we note that many studies, whether exploring advanced
techniques, new data, or both, rely on aggregation of data by some
arbitrary boundary. For example, Turner (2001) predicted gentrification
in the Washington, D.C. metro area by ranking census tracts in terms of
development. Chapple (2009) created a gentrification early warning
system by identifying low-income census tracts in central city
locations. Pollack, Bluestone, \& Billingham (2010) analyzed 42 census
block groups near rail stations in 12 metro areas across the United
States, studying changes between 1990 and 2000 for neighborhood
socioeconomic and housing characteristics. All of these studies, and
many more, relied on the aggregation of data at the census-tract or
census-block level. In contrast, this paper compares
boundary-aggregation techniques (specifically, aggregating by zip codes)
to a boundary-agnostic spatial lag technique and finds the latter to
outperform.

\hypertarget{data-and-methodology}{%
\section{Data and Methodology}\label{data-and-methodology}}

\hypertarget{methodology-overview}{%
\subsection{Methodology Overview}\label{methodology-overview}}

Our goal was to compare \emph{spatially-conscious} machine learning
predictive models to traditional feature engineering techniques. To
accomplish this comparison, we created three separate modeling datasets:

\begin{itemize}
\tightlist
\item
  \textbf{Base modeling data:} includes building characteristics such as
  size, taxable value, usage, and others
\item
  \textbf{Zip code modeling data:} includes the base data as well as
  aggregations of data at the zip code level
\item
  \textbf{Spatial lag modeling data:} includes the base data as well as
  aggregations of data within 500-meters of each building
\end{itemize}

\noindent The second and third modeling datasets are incremental
variations of the first, using competing feature engineering techniques
to extract additional predictive power from the data. We combined three
open-source data repositories provided by New York City via
\url{nyc.gov} and \url{data.cityofnewyork.us}. Our base modeling dataset
included all building records and associated sales information from
2003-2017. For each of the three modeling datasets, we also compared two
predictive modeling tasks, using a different dependent variable for
each:

\begin{enumerate}
\def\labelenumi{\arabic{enumi})}
\tightlist
\item
  \textbf{Classification task: probability of sale} The probability that
  a given property will sell in a given year (0,1)
\item
  \textbf{Regression task: sale-price-per-square-foot} Given that a
  property sells, how much is the sale-price-per-square-foot? (\$/SF)
\end{enumerate}

\noindent Table \ref{tab:modeltable} shows the six distinct modeling
task/data combinations.

\begin{table}[t]

\caption{\label{tab:model table}\label{tab:modeltable} Six Predictive Models}
\centering
\resizebox{\linewidth}{!}{
\begin{tabular}{r|l|l|l|l|l|l}
\hline
\# & Model & Model Task & Data & Outcome Var & Outcome Type & Eval Metric\\
\hline
1 & Probability of Sale & Classification & Base & Building Sold & Binary & AUC\\
\hline
2 & Probability of Sale & Classification & Zip Code & Building Sold & Binary & AUC\\
\hline
3 & Probability of Sale & Classification & Spatial Lag & Building Sold & Binary & AUC\\
\hline
4 & Sale Price & Regression & Base & Sale-Price-per-SF & Continuous & RMSE\\
\hline
5 & Sale Price & Regression & Zip Code & Sale-Price-per-SF & Continuous & RMSE\\
\hline
6 & Sale Price & Regression & Spatial Lag & Sale-Price-per-SF & Continuous & RMSE\\
\hline
\end{tabular}}
\end{table}

We conducted our analysis in a two-stage process. In Stage 1, we used the Random Forest algorithm to evaluate the suitability of the data for
our feature engineering assumptions. In Stage 2, we created subsets of
the modeling data based on the analysis conducted in Stage 1. We then
compared the performance of different algorithms across all modeling
datasets and prediction tasks. The following is an outline of our
complete analysis process:\newline

\noindent \textbf{Stage 1: Random Forest algorithm using all data}

\begin{enumerate}
\def\labelenumi{\arabic{enumi})}
\tightlist
\item
  Create a base modeling dataset by sourcing and combining building
  characteristic and sales data from open-source New York City
  repositories
\item
  Create a zip code modeling dataset by aggregating the base data at a
  zip code level and appending these features to the base data
\item
  Create a spatial lag modeling dataset by aggregating the base data
  within 500 meters of each building and appending these features to the
  base data
\item
  Train a Random Forest model on all three datasets, for both
  classification (probability of sale) and regression (sale price) tasks
\item
  Evaluate the performance of the various Random Forest models on
  hold-out test data
\item
  Analyze the prediction results by building type and geography,
  identifying those buildings for which our feature-engineering
  assumptions (e.g., 500-meter radii spatial lags) are most
  appropriate\newline
\end{enumerate}

\noindent \textbf{Stage 2: Many algorithms using refined data}

\begin{enumerate}
\def\labelenumi{\arabic{enumi})}
\setcounter{enumi}{6}
\tightlist
\item
  Create subsets of the modeling data based on analysis conducted in
  Stage 1
\item
  Train machine learning models on the refined modeling datasets using
  several algorithms, for both classification and regression tasks
\item
  Evaluate the performance of the various models on hold-out test data
\item
  Analyze the prediction results of the various algorithm/data/task
  combinations
\end{enumerate}

\hypertarget{data}{%
\subsection{Data}\label{data}}

\hypertarget{data-sources}{%
\subsubsection{Data Sources}\label{data-sources}}

The New York City government makes available an annual dataset which
describes all tax lots in the five boroughs. The Primary Land Use and
Tax Lot Output dataset, known as
\href{https://www1.nyc.gov/site/planning/data-maps/open-data/bytes-archive.page?sorts\%5Byear\%5D=0}{PLUTO}\footnote{https://www1.nyc.gov/site/planning/data-maps/open-data/bytes-archive.page?sorts{[}year{]}=0},
contains a single record for every tax lot in the city along with a
number of building-related and tax-related attributes such as year
built, assessed value, square footage, number of stories, and many more.
At the time of this writing, NYC had made this dataset available for all
years between 2002-2017, excluding 2008. For convenience, we also
exclude the 2002 dataset from our analysis because corresponding sales
information is not available for that year. Importantly for our
analysis, the latitude and longitude of the tax lots are also made
available, allowing us to locate in space each building and to build
geospatial features from the data.

Ultimately, we were interested in both the occurrence and the amount of
real estate sales transactions. Sales transactions are made available
separately by the New York City government, known as the
\href{http://www1.nyc.gov/site/finance/taxes/property-annualized-sales-update.page}{NYC
Rolling Sales Data}\footnote{http://www1.nyc.gov/site/finance/taxes/property-annualized-sales-update.page}.
At the time of this writing, sales transactions were available for the
years 2003-2017. The sales transactions data contains additional data
fields describing time, place, and amount of sale as well as additional
building characteristics. Crucially, the sales transaction data does not
include geographical coordinates, making it impossible to perform
geospatial analysis without first mapping the sales data to PLUTO.

\begin{figure}
\centering
\includegraphics{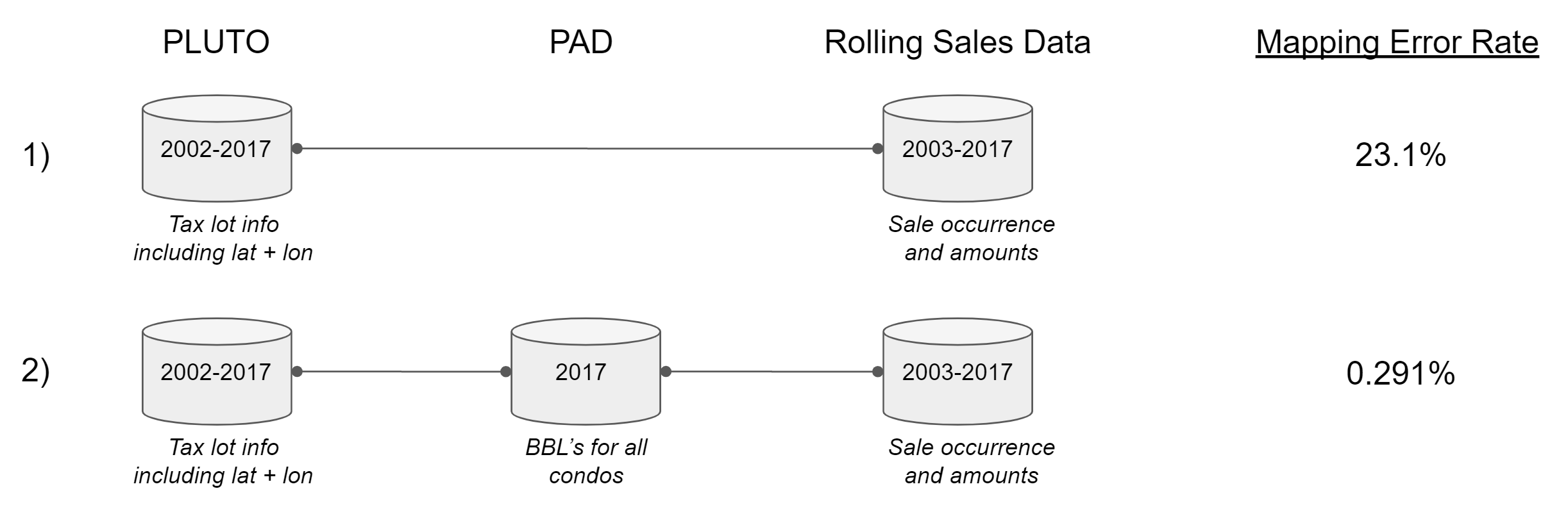}
\caption{\label{fig:Data Schema}Overview of Data Sources}
\end{figure}

Prior to mapping to PLUTO, we first had to transform the sales data to
include the proper mapping key. New York City uses a standard key of
Borough-Block-Lot to identify tax lots in the data. For example, 31 West
27th Street is located in Manhattan, on block 829 and lot 16; therefore,
its Borough-Block-Lot (BBL) is 1\_829\_16 (the 1 represents Manhattan).
The sales data contain BBL's at the building level; however, the sales
transactions data does not appropriately designate condos as their own
BBL's. Mapping the sales data directly to the PLUTO data results in a
mapping error rate of 23.1\% (mainly due to condos). Therefore, the
sales transactions data must first be mapped to another data source, the
NYC Property Address Directory, or
\href{https://data.cityofnewyork.us/City-Government/Property-Address-Directory/bc8t-ecyu/data}{PAD}\footnote{https://data.cityofnewyork.us/City-Government/Property-Address-Directory/bc8t-ecyu/data},
which contains an exhaustive list of all BBL's in NYC. After combining
the sales data with PAD, the data can then be mapped to PLUTO with an
error rate of 0.291\% (See: Figure \ref{fig:Data Schema}).

After combining the Sales Transactions data with PAD and PLUTO, we
filtered the resulting data for BBL's with less than or equal to 1
transaction per year. The final dataset is an exhaustive list of all tax
lots in NYC for every year between 2003-2017, whether that building was
sold, for what amount, and several other additional variables. A
description of all variables can be seen in Table
\ref{tab:descripTable}.

\begin{table}[t]

\caption{\label{tab:descripTable}\label{tab:descripTable} Description of Base Data}
\centering
\resizebox{\linewidth}{!}{
\begin{tabular}{l|l|l|l|l|l|l|l|l|l}
\hline
variable & type & nobs & mean & sd & mode & min & max & median & n\_missing\\
\hline
Annual\_Sales & Numeric & 12,012,780 & 2 & 8 & NA & 1 & 2,591 & 1 & 11,208,593\\
\hline
AssessLand & Numeric & 12,012,780 & 93,493 & 2,870,654 & 103,050 & 0 & 2,146,387,500 & 10,348 & 65\\
\hline
AssessTot & Numeric & 12,012,780 & 302,375 & 4,816,339 & 581,400 & 0 & 2,146,387,500 & 25,159 & 1,703,150\\
\hline
BldgArea & Numeric & 12,012,780 & 6,228 & 70,161 & 18,965 & 0 & 49,547,830 & 2,050 & 45\\
\hline
BldgDepth & Numeric & 12,012,780 & 46 & 34 & 50 & 0 & 9,388 & 42 & 44\\
\hline
BldgFront & Numeric & 12,012,780 & 25 & 33 & 100 & 0 & 9,702 & 20 & 44\\
\hline
Block & Numeric & 12,012,780 & 5,297 & 3,695 & 1 & 0 & 71,724 & 4,799 & 44\\
\hline
BoroCode & Numeric & 12,012,780 & 3 & 1 & 5 & 1 & 5 & 4 & 47\\
\hline
BsmtCode & Numeric & 12,012,780 & 2 & 2 & 0 & 0 & 3,213 & 2 & 859,406\\
\hline
BuiltFAR & Numeric & 12,012,780 & 1 & 10 & 3 & 0 & 8,695 & 1 & 850,554\\
\hline
ComArea & Numeric & 12,012,780 & 2,160 & 58,192 & 18,965 & 0 & 27,600,000 & 0 & 44\\
\hline
CommFAR & Numeric & 12,012,780 & 0 & 1 & 3 & 0 & 15 & 0 & 7,716,603\\
\hline
CondoNo & Numeric & 12,012,780 & 8 & 126 & 0 & 0 & 30,000 & 0 & 1,703,113\\
\hline
Easements & Numeric & 12,012,780 & 0 & 2 & 0 & 0 & 7,500 & 0 & 48\\
\hline
ExemptLand & Numeric & 12,012,780 & 37,073 & 2,718,194 & 0 & 0 & 2,146,387,500 & 1,290 & 65\\
\hline
ExemptTot & Numeric & 12,012,780 & 107,941 & 3,522,172 & 0 & 0 & 2,146,387,500 & 1,360 & 1,703,149\\
\hline
FacilFAR & Numeric & 12,012,780 & 2 & 2 & 5 & 0 & 15 & 2 & 7,716,603\\
\hline
FactryArea & Numeric & 12,012,780 & 126 & 3,890 & 0 & 0 & 1,324,592 & 0 & 850,555\\
\hline
GarageArea & Numeric & 12,012,780 & 130 & 5,154 & 0 & 0 & 2,677,430 & 0 & 850,554\\
\hline
GROSS SQUARE FEET & Numeric & 12,012,780 & 4,423 & 45,691 & NA & 0 & 14,962,152 & 1,920 & 11,217,669\\
\hline
lat & Numeric & 12,012,780 & 41 & 0 & 41 & 40 & 41 & 41 & 427,076\\
\hline
lon & Numeric & 12,012,780 & -74 & 0 & -74 & -78 & -74 & -74 & 427,076\\
\hline
Lot & Numeric & 12,012,780 & 115 & 655 & 10 & 0 & 9,999 & 38 & 44\\
\hline
LotArea & Numeric & 12,012,780 & 7,852 & 362,618 & 5,716 & 0 & 214,755,710 & 2,514 & 44\\
\hline
LotDepth & Numeric & 12,012,780 & 104 & 69 & 84 & 0 & 9,999 & 100 & 45\\
\hline
LotFront & Numeric & 12,012,780 & 40 & 74 & 113 & 0 & 9,999 & 25 & 44\\
\hline
LotType & Numeric & 12,012,780 & 5 & 1 & 5 & 0 & 9 & 5 & 865,340\\
\hline
NumBldgs & Numeric & 12,012,780 & 1 & 4 & 1 & 0 & 2,740 & 1 & 46\\
\hline
NumFloors & Numeric & 12,012,780 & 2 & 2 & 4 & 0 & 300 & 2 & 44\\
\hline
OfficeArea & Numeric & 12,012,780 & 742 & 21,566 & 0 & 0 & 5,009,319 & 0 & 850,556\\
\hline
OtherArea & Numeric & 12,012,780 & 673 & 49,848 & 0 & 0 & 27,600,000 & 0 & 850,555\\
\hline
ProxCode & Numeric & 12,012,780 & 1 & 2 & 1 & 0 & 5,469 & 1 & 197,927\\
\hline
ResArea & Numeric & 12,012,780 & 3,921 & 31,882 & 0 & 0 & 35,485,021 & 1,776 & 44\\
\hline
ResidFAR & Numeric & 12,012,780 & 1 & 1 & 2 & 0 & 12 & 1 & 7,716,603\\
\hline
RetailArea & Numeric & 12,012,780 & 309 & 14,394 & 6,965 & 0 & 21,999,988 & 0 & 850,554\\
\hline
SALE PRICE & Numeric & 12,012,780 & 884,036 & 13,757,706 & NA & 0 & 4,111,111,766 & 319,000 & 11,208,593\\
\hline
sale\_psf & Numeric & 12,012,780 & 220 & 5,153 & NA & 0 & 1,497,500 & 114 & 11,250,396\\
\hline
SALE\_YEAR & Numeric & 12,012,780 & 2,009 & 5 & NA & 2,003 & 2,017 & 2,009 & 11,208,593\\
\hline
Sold & Numeric & 12,012,780 & 0 & 0 & 0 & 0 & 1 & 0 & 0\\
\hline
StrgeArea & Numeric & 12,012,780 & 169 & 5,810 & 12,000 & 0 & 1,835,150 & 0 & 850,554\\
\hline
TOTAL\_SALES & Numeric & 12,012,780 & 884,036 & 13,757,706 & NA & 0 & 4,111,111,766 & 319,000 & 11,208,593\\
\hline
UnitsRes & Numeric & 12,012,780 & 4 & 36 & 0 & 0 & 20,811 & 1 & 45\\
\hline
UnitsTotal & Numeric & 12,012,780 & 4 & 42 & 1 & 0 & 44,276 & 2 & 47\\
\hline
Year & Numeric & 12,012,780 & 2,010 & 4 & 2,017 & 2,003 & 2,017 & 2,011 & 0\\
\hline
YearAlter1 & Numeric & 12,012,780 & 159 & 540 & 2,000 & 0 & 2,017 & 0 & 45\\
\hline
YearAlter2 & Numeric & 12,012,780 & 20 & 202 & 0 & 0 & 2,017 & 0 & 48\\
\hline
YearBuilt & Numeric & 12,012,780 & 1,830 & 449 & 1,884 & 0 & 2,040 & 1,930 & 47\\
\hline
ZipCode & Numeric & 12,012,780 & 11,007 & 537 & 10,301 & 0 & 11,697 & 11,221 & 59,956\\
\hline
Address & Character & 12,012,780 & NA & NA & NA & NA & NA & NA & 17,902\\
\hline
AssessTotal & Character & 12,012,780 & NA & NA & NA & NA & NA & NA & 10,309,712\\
\hline
bbl & Character & 12,012,780 & NA & NA & NA & NA & NA & NA & 0\\
\hline
BldgClass & Character & 12,012,780 & NA & NA & NA & NA & NA & NA & 16,372\\
\hline
Borough & Character & 12,012,780 & NA & NA & NA & NA & NA & NA & 0\\
\hline
BUILDING CLASS AT PRESENT & Character & 12,012,780 & NA & NA & NA & NA & NA & NA & 11,219,514\\
\hline
BUILDING CLASS AT TIME OF SALE & Character & 12,012,780 & NA & NA & NA & NA & NA & NA & 11,208,593\\
\hline
BUILDING CLASS CATEGORY & Character & 12,012,780 & NA & NA & NA & NA & NA & NA & 11,208,765\\
\hline
Building\_Type & Character & 12,012,780 & NA & NA & NA & NA & NA & NA & 16,372\\
\hline
CornerLot & Character & 12,012,780 & NA & NA & NA & NA & NA & NA & 11,163,751\\
\hline
ExemptTotal & Character & 12,012,780 & NA & NA & NA & NA & NA & NA & 10,309,712\\
\hline
FAR & Character & 12,012,780 & NA & NA & NA & NA & NA & NA & 11,162,270\\
\hline
IrrLotCode & Character & 12,012,780 & NA & NA & NA & NA & NA & NA & 16,310\\
\hline
MaxAllwFAR & Character & 12,012,780 & NA & NA & NA & NA & NA & NA & 4,296,221\\
\hline
OwnerName & Character & 12,012,780 & NA & NA & NA & NA & NA & NA & 137,048\\
\hline
OwnerType & Character & 12,012,780 & NA & NA & NA & NA & NA & NA & 10,445,328\\
\hline
TAX CLASS AT PRESENT & Character & 12,012,780 & NA & NA & NA & NA & NA & NA & 11,219,514\\
\hline
TAX CLASS AT TIME OF SALE & Character & 12,012,780 & NA & NA & NA & NA & NA & NA & 11,208,593\\
\hline
ZoneDist1 & Character & 12,012,780 & NA & NA & NA & NA & NA & NA & 18,970\\
\hline
ZoneDist2 & Character & 12,012,780 & NA & NA & NA & NA & NA & NA & 11,715,653\\
\hline
\end{tabular}}
\end{table}

\hypertarget{global-filtering-of-the-data}{%
\subsubsection{Global Filtering of the
Data}\label{global-filtering-of-the-data}}

We only included building categories of significant interest in our
initial modeling data. Generally speaking, by significant interest we
are referring to building types that are regularly bought and sold on
the free market. These include residences, office buildings, and
industrial buildings, and exclude things like government-owned buildings
and hospitals. We also excluded hotels as they tend to be comparatively
rare in the data and exhibit unique sales characteristics. The included
building types are displayed in Table \ref{tab:categoryTable}.

\begin{table}[t]

\caption{\label{tab:unnamed-chunk-6}\label{tab:categoryTable} Included Building Cateogory Codes}
\centering
\resizebox{\linewidth}{!}{
\begin{tabular}{l|l}
\hline
Category & Description\\
\hline
A & ONE FAMILY DWELLINGS\\
\hline
B & TWO FAMILY DWELLINGS\\
\hline
C & WALK UP APARTMENTS\\
\hline
D & ELEVATOR APARTMENTS\\
\hline
F & FACTORY AND INDUSTRIAL BUILDINGS\\
\hline
G & GARAGES AND GASOLINE STATIONS\\
\hline
L & LOFT BUILDINGS\\
\hline
O & OFFICES\\
\hline
\end{tabular}}
\end{table}

The data were further filtered to include only records with equal to or
less than 2 buildings per tax lot, effectively excluding large outliers
in the data such as the World Trade Center and Stuyvesant Town. The
global filtering of the dataset reduced the base modeling data from
12,012,780 records down to 8,247,499, retaining 68.6\% of the original
data.

\hypertarget{exploratory-data-analysis}{%
\subsubsection{Exploratory Data
Analysis}\label{exploratory-data-analysis}}

The data contain building and sale records across the five boroughs of
New York City for the years 2003-2017. One challenge with creating a
predictive model of real estate sales data is the heterogeneity within
the data in terms of frequency of sales and sale price. These two
metrics (sale occurrence and amount) vary meaningfully across year,
borough and building class (among other attributes). Table
\ref{tab:by_year} displays statistics which describe the base dataset
(pre-filtered) by year. Note how the frequency of transactions (\# of
Sales) and the sale amount (Median Sale \$/SF) tend to covary,
particularly through the downturn of 2009-2012. This covariance may be
due to the fact that the relative size of transactions tends to decrease
as capital becomes more constrained.

\begin{table}[t]

\caption{\label{tab:by_year}\label{tab:by_year} Sales By Year}
\centering
\resizebox{\linewidth}{!}{
\begin{tabular}{r|r|r|l|l}
\hline
Year & N & \# Sales & Median Sale & Median Sale \$/SF\\
\hline
2003 & 850515 & 78919 & \$218,000 & \$79.37\\
\hline
2004 & 852563 & 81794 & \$292,000 & \$124.05\\
\hline
2005 & 854862 & 77815 & \$360,500 & \$157.76\\
\hline
2006 & 857473 & 70928 & \$400,000 & \$168.07\\
\hline
2007 & 860480 & 61880 & \$385,000 & \$139.05\\
\hline
2009 & 860519 & 43304 & \$245,000 & \$41.25\\
\hline
2010 & 860541 & 41826 & \$273,000 & \$75.35\\
\hline
2011 & 860320 & 40852 & \$263,333 & \$56.99\\
\hline
2012 & 859329 & 47036 & \$270,708 & \$52.72\\
\hline
2013 & 859372 & 50408 & \$315,000 & \$89.44\\
\hline
2014 & 858914 & 51386 & \$350,000 & \$115.71\\
\hline
2015 & 859464 & 53208 & \$375,000 & \$135.62\\
\hline
2016 & 859205 & 53772 & \$385,530 & \$147.06\\
\hline
2017 & 859223 & 51059 & \$430,000 & \$171.71\\
\hline
\end{tabular}}
\end{table}

We observe similar variances across asset types. Table
\ref{tab:by_class} shows all buildings classes in the 2003-2017 period.
Unsurprisingly, residences tend to have the highest volume of sales
while offices tend to have the highest sale prices.

\begin{table}[t]

\caption{\label{tab:by_class}\label{tab:by_class} Sales By Asset Class}
\centering
\resizebox{\linewidth}{!}{
\begin{tabular}{l|l|r|r|l|l}
\hline
Bldg Code & Build Type & N & \# Sales & Median Sale & Median Sale \$/SF\\
\hline
A & One Family Dwellings & 4435615 & 252283 & \$320,000 & \$215.85\\
\hline
B & Two Family Dwellings & 3431762 & 219492 & \$340,000 & \$155.79\\
\hline
C & Walk Up Apartments & 1873447 & 135203 & \$330,000 & \$67.20\\
\hline
D & Elevator Apartments & 188689 & 45635 & \$398,000 & \$4.69\\
\hline
E & Warehouses & 84605 & 5126 & \$200,000 & \$31.48\\
\hline
F & Factory & 67174 & 4440 & \$350,000 & \$56.44\\
\hline
G & Garages & 221620 & 13965 & \$0 & \$78.57\\
\hline
H & Hotels & 10807 & 619 & \$5,189,884 & \$184.82\\
\hline
I & Hospitals & 17650 & 687 & \$600,000 & \$62.66\\
\hline
J & Theatres & 2662 & 152 & \$113,425 & \$4.01\\
\hline
K & Retail & 265101 & 14841 & \$200,000 & \$60.63\\
\hline
L & Loft & 18239 & 1259 & \$1,937,500 & \$101.36\\
\hline
M & Religious & 78063 & 1320 & \$375,000 & \$91.78\\
\hline
N & Asylum & 8498 & 190 & \$275,600 & \$35.90\\
\hline
O & Office & 93973 & 5294 & \$550,000 & \$143.29\\
\hline
P & Public Assembly & 15292 & 437 & \$350,000 & \$85.47\\
\hline
Q & Recreation & 55193 & 232 & \$0 & \$0\\
\hline
R & Condo & 78188 & 40157 & \$444,750 & \$12.65\\
\hline
S & Mixed Use Residence & 467555 & 29396 & \$250,000 & \$78.29\\
\hline
T & Transportation & 4012 & 49 & \$0 & \$0\\
\hline
U & Utility & 32802 & 129 & \$0 & \$175\\
\hline
V & Vacant & 449667 & 29091 & \$0 & \$134.70\\
\hline
W & Educational & 38993 & 704 & \$0 & \$0\\
\hline
Y & Gov't & 7216 & 44 & \$21,451.50 & \$0.30\\
\hline
Z & Misc & 49583 & 2740 & \$0 & \$0\\
\hline
\end{tabular}}
\end{table}

Sale-price-per-square-foot, in particular, varies considerably across
geography and asset class. Table \ref{tab:by_class_boro} shows the
breakdown of sales prices by borough and asset class. Manhattan tends to
command the highest sale-price-per-square-foot across asset types.
``Commercial'' asset types such as Office and Elevator Apartments tend
to fetch much lower price-per-square-foot than do residential classes
such as one and two-family dwellings. Table \ref{tab:by_class_boro_num}
shows the number of transactions across the same dimensions.

\begin{table}[t]

\caption{\label{tab:by_class_boro}\label{tab:by_class_boro} Sale Price Per Square Foot by Asset Class and Borough}
\centering
\resizebox{\linewidth}{!}{
\begin{tabular}{l|l|l|l|l|l}
\hline
Build Type & BK & BX & MN & QN & SI\\
\hline
Elevator Apartments & \$2.65 & \$1.74 & \$10.80 & \$1.87 & \$1.23\\
\hline
Factory & \$33.33 & \$53.19 & \$135.62 & \$92.42 & \$55.01\\
\hline
Garages & \$78.94 & \$80.57 & \$94.43 & \$71.11 & \$67.46\\
\hline
Loft & \$46.32 & \$78.26 & \$141.56 & \$150.37 & \$61.82\\
\hline
Office & \$118.52 & \$123.04 & \$225.96 & \$148.45 & \$105\\
\hline
One Family Dwellings & \$221.26 & \$176.98 & \$757.58 & \$232.69 & \$203.88\\
\hline
Two Family Dwellings & \$140.95 & \$131.06 & \$296.10 & \$181.84 & \$160.76\\
\hline
Walk Up Apartments & \$69.97 & \$84.05 & \$50.61 & \$36.94 & \$75.38\\
\hline
\end{tabular}}
\end{table}

\begin{table}[t]

\caption{\label{tab:by_class_boro_num}\label{tab:by_class_boro_num} Number of Sales by Asset Class and Borough}
\centering
\resizebox{\linewidth}{!}{
\begin{tabular}{l|l|l|l|l|l}
\hline
Build Type & BK & BX & MN & QN & SI\\
\hline
Elevator Apartments & 8,377 & 4,252 & 23,641 & 9,196 & 169\\
\hline
Factory & 2,265 & 453 & 109 & 1,520 & 93\\
\hline
Garages & 5,386 & 2,659 & 1,097 & 4,000 & 823\\
\hline
Loft & 119 & 21 & 1,108 & 8 & 3\\
\hline
Office & 1,112 & 340 & 2,081 & 1,162 & 599\\
\hline
One Family Dwellings & 45,009 & 17,508 & 1,654 & 126,333 & 61,779\\
\hline
Two Family Dwellings & 83,547 & 25,920 & 1,566 & 83,940 & 24,519\\
\hline
Walk Up Apartments & 63,552 & 18,075 & 19,824 & 31,932 & 1,820\\
\hline
\end{tabular}}
\end{table}

\hypertarget{feature-engineering}{%
\subsection{Feature Engineering}\label{feature-engineering}}

\hypertarget{base-modeling-data}{%
\subsubsection{Base Modeling Data}\label{base-modeling-data}}

We constructed the base modeling dataset by combining several
open-source data repositories, outlined in the Data Sources section. In
addition to the data provided by New York City, several additional
features were engineered and appended to the base data. A summary table
of the additional features is presented in Table
\ref{tab:baseModelDataFeats}. A binary variable was created to indicate
whether a tax lot had a building on it (i.e., whether it was an empty
plot of land). In addition, building types were quantified by what
percent of their square footage belonged to the major property types:
Commercial, Residential, Office, Retail, Garage, Storage, Factory and
Other.

\begin{table}[t]

\caption{\label{tab:Table 1}\label{tab:baseModelDataFeats} Base Modeling Data Features}
\centering
\resizebox{\linewidth}{!}{
\begin{tabular}{l|l|l|l|l}
\hline
Feature & Min & Median & Mean & Max\\
\hline
has\_building\_area & 0 & 1.00 & 1.00 & 1.00\\
\hline
Percent\_Com & 0 & 0.00 & 0.16 & 1.00\\
\hline
Percent\_Res & 0 & 1.00 & 0.82 & 1.00\\
\hline
Percent\_Office & 0 & 0.00 & 0.07 & 1.00\\
\hline
Percent\_Retail & 0 & 0.00 & 0.04 & 1.00\\
\hline
Percent\_Garage & 0 & 0.00 & 0.01 & 1.00\\
\hline
Percent\_Storage & 0 & 0.00 & 0.02 & 1.00\\
\hline
Percent\_Factory & 0 & 0.00 & 0.00 & 1.00\\
\hline
Percent\_Other & 0 & 0.00 & 0.00 & 1.00\\
\hline
Last\_Sale\_Price & 0 & 312.68 & 531.02 & 62,055.59\\
\hline
Last\_Sale\_Price\_Total & 2 & 2,966,835.00 & 12,844,252.00 & 1,932,900,000.00\\
\hline
Years\_Since\_Last\_Sale & 1 & 4.00 & 5.05 & 14.00\\
\hline
SMA\_Price\_2\_year & 0 & 296.92 & 500.89 & 62,055.59\\
\hline
SMA\_Price\_3\_year & 0 & 294.94 & 495.29 & 62,055.59\\
\hline
SMA\_Price\_5\_year & 0 & 300.12 & 498.82 & 62,055.59\\
\hline
Percent\_Change\_SMA\_2 & -1 & 0.00 & 685.69 & 15,749,999.50\\
\hline
Percent\_Change\_SMA\_5 & -1 & 0.00 & 337.77 & 6,299,999.80\\
\hline
EMA\_Price\_2\_year & 0 & 288.01 & 482.69 & 62,055.59\\
\hline
EMA\_Price\_3\_year & 0 & 283.23 & 471.98 & 62,055.59\\
\hline
EMA\_Price\_5\_year & 0 & 278.67 & 454.15 & 62,055.59\\
\hline
Percent\_Change\_EMA\_2 & -1 & 0.00 & 422.50 & 9,415,128.85\\
\hline
Percent\_Change\_EMA\_5 & -1 & 0.06 & 308.05 & 5,341,901.60\\
\hline
\end{tabular}}
\end{table}

Importantly, we created two variables from the sale prices: A
price-per-square-foot figure (Sale\_Price) and a total sale price
(Sale\_Price\_Total). Sale-price-per-square-foot eventually became the
outcome variable in the regression modeling tasks. We then created a
feature to carry forward the previous sale price of a tax lot, if there
was one, through successive years. The previous sale price was then used
to create simple moving averages (SMA), exponential moving averages
(EMA), and percent change measurements between the moving averages. In
total, 69 variables were input to the feature engineering process, and
92 variables were output. The final base modeling dataset was 92
variables by 8,247,499 rows.

\hypertarget{zip-code-modeling-data}{%
\subsubsection{Zip Code Modeling Data}\label{zip-code-modeling-data}}

The first of the two comparative modeling datasets was the zip code
modeling data. We aggregated the base data at a zip code level and then
generated several features to describe the characteristics of where each
tax lot resides. A summary table of the zip code level features is
presented in \ref{tab:zipcodemodelfeats}.

\begin{table}[t]

\caption{\label{tab:Table 2}\label{tab:zipcodemodelfeats} Zip Code Modeling Data Features}
\centering
\resizebox{\linewidth}{!}{
\begin{tabular}{l|l|l|l|l}
\hline
Feature & Min & Median & Mean & Max\\
\hline
Last Year Zip Sold & 0.00 & 27.00 & 31.14 & 112.00\\
\hline
Last Year Zip Sold Percent Ch & -1.00 & 0.00 &  & \\
\hline
Last Sale Price zip code average & 0.00 & 440.95 & 522.87 & 1,961.21\\
\hline
Last Sale Price Total zip code average & 10.00 & 5,312,874.67 & 11,877,688.55 & 1,246,450,000.00\\
\hline
Last Sale Date zip code average & 12,066.00 & 13,338.21 & 13,484.39 & 17,149.00\\
\hline
Years Since Last Sale zip code average & 1.00 & 4.84 & 4.26 & 11.00\\
\hline
SMA Price 2 year zip code average & 34.31 & 429.26 & 501.15 & 2,092.41\\
\hline
SMA Price 3 year zip code average & 34.31 & 422.04 & 496.47 & 2,090.36\\
\hline
SMA Price 5 year zip code average & 39.48 & 467.04 & 520.86 & 2,090.36\\
\hline
Percent Change SMA 2 zip code average & -0.20 & 0.04 & 616.47 & 169,999.90\\
\hline
Percent Change SMA 5 zip code average & -0.09 & 0.03 & 341.68 & 113,333.27\\
\hline
EMA Price 2 year zip code average & 30.77 & 401.43 & 479.38 & 1,883.81\\
\hline
EMA Price 3 year zip code average & 33.48 & 419.11 & 479.95 & 1,781.38\\
\hline
EMA Price 5 year zip code average & 29.85 & 431.89 & 472.80 & 1,506.46\\
\hline
Percent Change EMA 2 zip code average & -0.16 & 0.06 & 388.90 & 107,368.37\\
\hline
Percent Change EMA 5 zip code average & -0.08 & 0.07 & 326.17 & 107,368.38\\
\hline
Last Sale Price bt only & 0.00 & 357.71 & 485.97 & 6,401.01\\
\hline
Last Sale Price Total bt only & 10.00 & 3,797,461.46 & 11,745,130.56 & 1,246,450,000.00\\
\hline
Last Sale Date bt only & 12,055.00 & 13,331.92 & 13,497.75 & 17,149.00\\
\hline
Years Since Last Sale bt only & 1.00 & 4.78 & 4.30 & 14.00\\
\hline
SMA Price 2 year bt only & 0.00 & 347.59 & 462.67 & 5,519.39\\
\hline
SMA Price 3 year bt only & 0.00 & 345.40 & 458.50 & 5,104.51\\
\hline
SMA Price 5 year bt only & 0.00 & 372.30 & 481.09 & 4,933.05\\
\hline
Percent Change SMA 2 bt only & -0.55 & 0.03 & 600.10 & 425,675.69\\
\hline
Percent Change SMA 5 bt only & -0.33 & 0.02 & 338.15 & 188,888.78\\
\hline
EMA Price 2 year bt only & 0.00 & 332.98 & 442.79 & 5,103.51\\
\hline
EMA Price 3 year bt only & 0.00 & 332.79 & 443.02 & 4,754.95\\
\hline
EMA Price 5 year bt only & 0.00 & 340.57 & 436.70 & 4,270.37\\
\hline
Percent Change EMA 2 bt only & -0.47 & 0.06 & 377.17 & 254,462.97\\
\hline
Percent Change EMA 5 bt only & -0.34 & 0.06 & 335.17 & 178,947.30\\
\hline
\end{tabular}}
\end{table}

The base model data features were aggregated to a zip code level and
appended, including the SMA, EMA and percent change calculations. We
then added another set of features, denoted as ``bt\_only,'' which again
aggregated the base features but only included tax lots of the same
building type. In total, the zip code feature engineering process input
92 variables and output 122 variables.

\hypertarget{spatial-lag-modeling-data}{%
\subsubsection{Spatial Lag Modeling
Data}\label{spatial-lag-modeling-data}}

Spatial lags are variables created from physically proximate
observations. For example, calculating the average age of all buildings
within 100 meters of a tax lot constitutes a spatial lag. Creating
spatial lags presents both advantages and disadvantages in the modeling
process. Spatial lags allow for much more fine-tuned measurements of a
building's surrounding area. Intuitively, knowing the average sale price
of all buildings within 500 meters of a building can be more informative
than knowing the sale prices of all buildings in the same zip code.
However, creating spatial lags is computationally expensive.
Additionally, it can be challenging to set a proper radius for the
spatial lag calculation; in a city, 500 meters may be appropriate (for
specific building types), whereas several kilometers or more may be
appropriate for less densely populated areas. In this paper, we present
a solution for the computational challenges and suggest a potential
approach to solving the radius-choice problem.

\hypertarget{creating-the-point-neighbor-relational-graph}{%
\paragraph{Creating the Point-Neighbor Relational
Graph}\label{creating-the-point-neighbor-relational-graph}}

\begin{figure}
\centering
\includegraphics{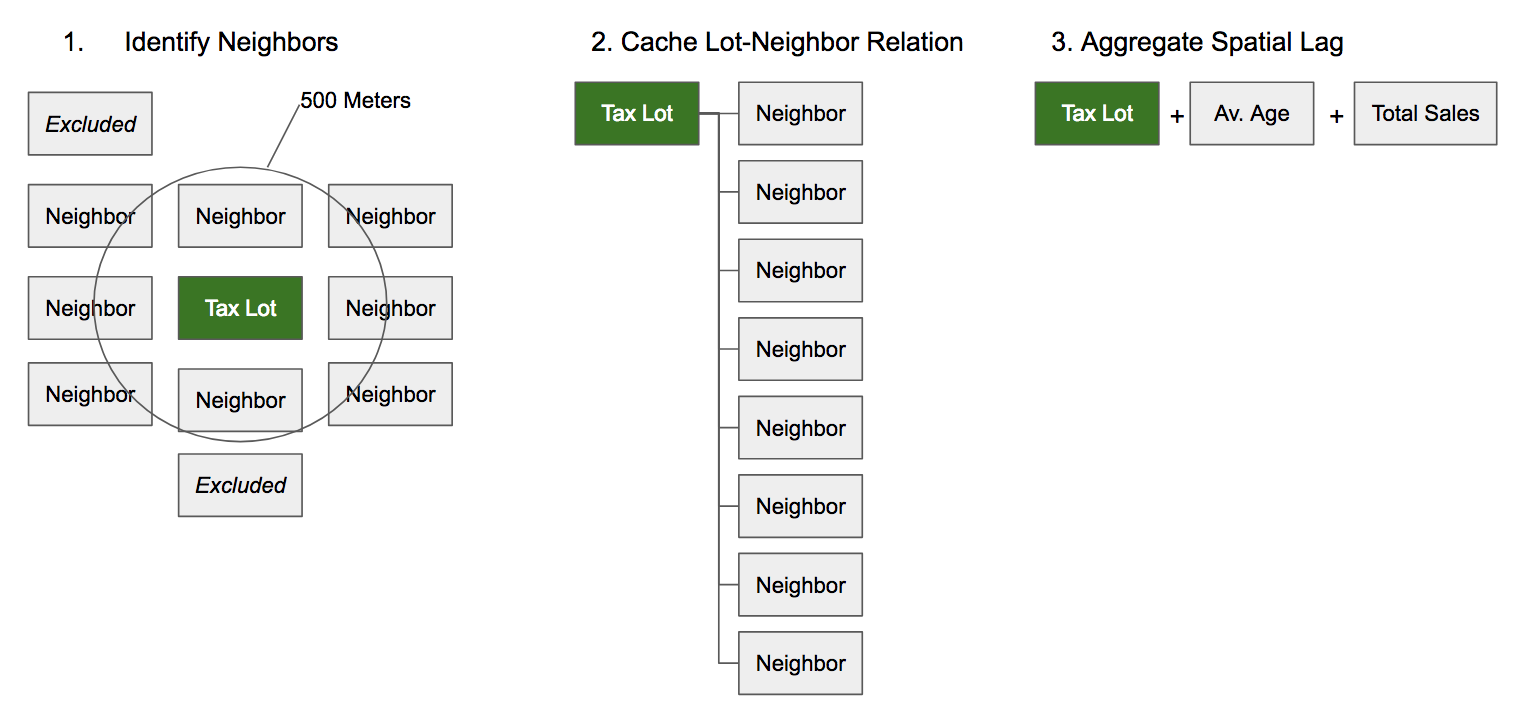}
\caption{\label{fig:Spatial Lag Feataure Process}Spatial Lag Feature
Creation Process}
\end{figure}

To build our spatial lags, for each point in the data, we must identify
which of all other points in the data fall within a specified radius.
This neighbor identification process requires iteratively running
point-in-polygon operations. This process is conceptually illustrated in
figure \ref{fig:Spatial Lag Feataure Process}.

Given that, for every point \(q_i\) in our dataset, we need to determine
whether every other point \(q_i\) falls within a given radius, this
means that we can approximate the time-complexity of our operation as:

\[
O(N(N-1))
\]

Since the number of operations approaches \(N^2\), calculating spatial
lags for all 8,247,499 observations in our modeling data would be
infeasible from a time and computation perspective. Assuming that tax
lots rarely if ever move over time, we first reduced the task to the
number of unique tax lots in New York City from 2003-2017, which is
514,124 points. Next, we implemented an indexing technique that greatly
speeds up the process of creating a point-neighbor relational graph. The
indexing technique both reduces the relative search space for each
computation and also allows for parallelization of the point-in-polygon
operations by dividing the data into a gridded space. The gridded
spatial indexing process is outlined in Algorithm \ref{alg:spatial1}.

\begin{algorithm}
\caption{Gridded Spatial Indexing}
\label{alg:spatial1}
\begin{algorithmic}[1]
\For{\texttt{each grid partition $G$}}
\State \texttt{Extract all points points $G_i$ contained within partition $G$}
\State \texttt{Calculate convex hull $H(G)$ such that the buffer extends to distance $d$}
\State \texttt{Define Search space $S$ as all points within Convex hull $H(G)$}
\State \texttt{Extract all points $S_i$ contained within $S$}
\For{\texttt{each data point $G_i$}}
\State \texttt{Identify all points points in $S_i$ that fall within $abs(G_i+d)$}
\EndFor
\EndFor
\end{algorithmic}
\end{algorithm}

Each gridded partition of the data is married with a corresponding
search space \(S\), which is the convex hull of the partition space
buffered by the maximum distance \(d\). In our case, we buffered the
search space by 500 meters. Choosing an appropriate radius for buffering
presents an additional challenge in creating spatially-conscious machine
learning predictive models. In this paper, we chose an arbitrary radius,
and use a two-stage modeling process to test the appropriateness of that
assumption. Future work may want to explore implementing an adaptive
bandwidth technique using cross-validation to determine the optimal
radius for each property.

By partitioning the data into spatial grids, we were able to reduce the
search space for each operation by an arbitrary number of partitions
\(G\). This improves the base run-time complexity to:

\[
O(N(\frac{N-1}{G})
\]

\begin{figure}
\centering
\includegraphics{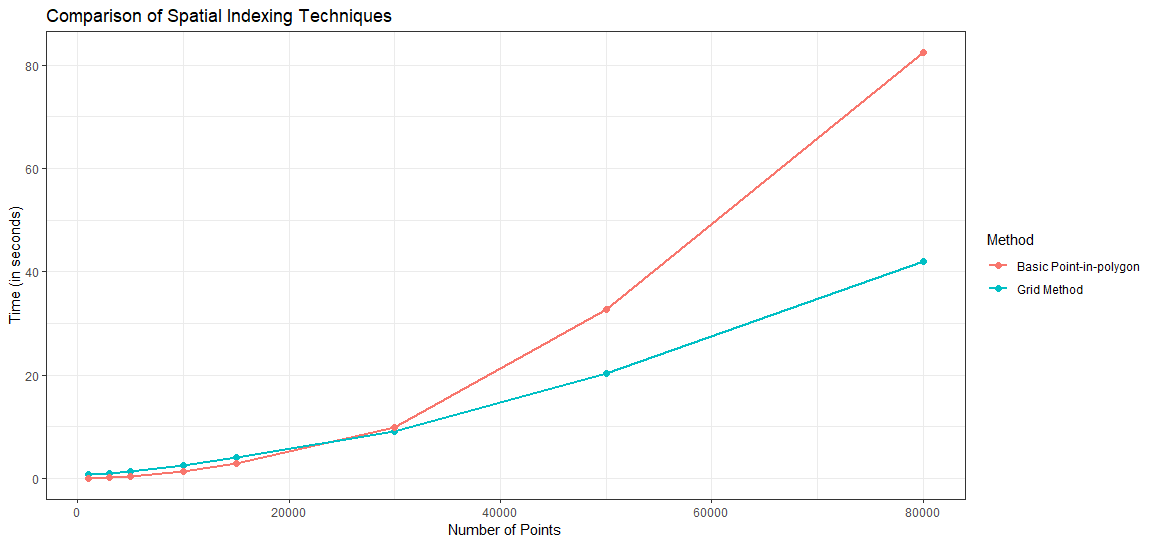}
\caption{\label{fig:Spatial Indexing Process}Spatial Index Time
Comparison}
\end{figure}

\noindent By making G arbitrarily large (bounded by computational
resources only), we reduced the runtime substantially. Furthermore,
binning the operations into grids allowed us to parallelize the
computation, further reducing the overall runtime. Figure
\ref{fig:Spatial Indexing Process} shows a comparison of computation
times between the basic point-in-polygon technique and a sequential
version of the grided indexing technique. Note that the grid method
starts as slower than the basic point-in-polygon technique due to
pre-processing overhead, but quickly wins out in terms of speed as the
complexity of the task increases. This graph also does not reflect the
parallelization of the grid method, which further reduced the time
required to calculate the point-neighbor relational graph.

\hypertarget{calculating-spatial-lags}{%
\paragraph{Calculating Spatial Lags}\label{calculating-spatial-lags}}

Once we constructed the point-neighbor relational graph, we then used
the graph to aggregate the data into spatial lag variables. One
advantage of using spatial lags is the abundant number of potential
features which can be engineered. Spatial lags can be weighted based on
a distance function, e.g., physically closer observations can be given
more weight. For our modeling purposes, we created two sets of features:
inverse-distance weighted features (denoted with a "\_dist" in Table
\ref{tab:SpLAgFeats}) and simple average features (denoted with
"\_basic" in Table \ref{tab:SpLAgFeats}).

Temporal and spatial derivatives of the spatial lag features, presented
in Table \ref{tab:SpLAgFeats}, were also added to the model, including:
variables weighted by Euclidean distance (``dist''), basic averages of
the spatial lag radius (``basic mean''), SMA for 2 years, 3 years and 5
years, EMA for 2 years, 3 years and 5 years, and year-over-year percent
changes for all variables (``perc change''). In total, the spatial lag
feature engineering process input 92 variables and output 194 variables.

\begin{table}[t]

\caption{\label{tab:SpLagFeats}\label{tab:SpLAgFeats} All Spatial Lag Features}
\centering
\resizebox{\linewidth}{!}{
\begin{tabular}{l|l|l|l|l}
\hline
Feature & Min & Median & Mean & Max\\
\hline
Radius\_Total\_Sold\_In\_Year & 1.00 & 20.00 & 24.00 & 201.00\\
\hline
Radius\_Average\_Years\_Since\_Last\_Sale & 1.00 & 4.43 & 4.27 & 14.00\\
\hline
Radius\_Res\_Units\_Sold\_In\_Year & 0.00 & 226.00 & 289.10 & 2,920.00\\
\hline
Radius\_All\_Units\_Sold\_In\_Year & 0.00 & 255.00 & 325.94 & 2,923.00\\
\hline
Radius\_SF\_Sold\_In\_Year & 0.00 & 259,403.00 & 430,891.57 & 8,603,639.00\\
\hline
Radius\_Total\_Sold\_In\_Year\_sum\_over\_2\_years & 2.00 & 41.00 & 48.15 & 256.00\\
\hline
Radius\_Average\_Years\_Since\_Last\_Sale\_sum\_over\_2\_years & 2.00 & 9.25 & 8.70 & 26.00\\
\hline
Radius\_Res\_Units\_Sold\_In\_Year\_sum\_over\_2\_years & 0.00 & 493.00 & 584.67 & 3,397.00\\
\hline
Radius\_All\_Units\_Sold\_In\_Year\_sum\_over\_2\_years & 1.00 & 555.00 & 660.67 & 4,265.00\\
\hline
Radius\_SF\_Sold\_In\_Year\_sum\_over\_2\_years & 2,917.00 & 580,947.00 & 872,816.44 & 14,036,469.00\\
\hline
Radius\_Total\_Sold\_In\_Year\_percent\_change & -0.99 & 0.00 & 0.27 & 77.00\\
\hline
Radius\_Average\_Years\_Since\_Last\_Sale\_percent\_change & -0.91 & 0.13 & 0.26 & 8.00\\
\hline
Radius\_Res\_Units\_Sold\_In\_Year\_percent\_change & -1.00 & -0.04 &  & \\
\hline
Radius\_All\_Units\_Sold\_In\_Year\_percent\_change & -1.00 & -0.04 &  & \\
\hline
Radius\_SF\_Sold\_In\_Year\_percent\_change & -1.00 & -0.02 &  & \\
\hline
Radius\_Total\_Sold\_In\_Year\_sum\_over\_2\_years\_percent\_change & -0.96 & -0.03 & 0.03 & 15.00\\
\hline
Radius\_Average\_Years\_Since\_Last\_Sale\_sum\_over\_2\_years\_percent\_change & -0.72 & 0.12 & 0.17 & 2.50\\
\hline
Radius\_Res\_Units\_Sold\_In\_Year\_sum\_over\_2\_years\_percent\_change & -1.00 & -0.04 &  & \\
\hline
Radius\_All\_Units\_Sold\_In\_Year\_sum\_over\_2\_years\_percent\_change & -0.99 & -0.04 & 0.12 & 84.00\\
\hline
Radius\_SF\_Sold\_In\_Year\_sum\_over\_2\_years\_percent\_change & -0.98 & -0.04 & 0.18 & 361.55\\
\hline
Percent\_Com\_dist & 0.00 & 0.04 & 0.07 & 0.56\\
\hline
Percent\_Res\_dist & 0.00 & 0.46 & 0.43 & 0.66\\
\hline
Percent\_Office\_dist & 0.00 & 0.01 & 0.03 & 0.48\\
\hline
Percent\_Retail\_dist & 0.00 & 0.02 & 0.02 & 0.09\\
\hline
Percent\_Garage\_dist & 0.00 & 0.00 & 0.00 & 0.27\\
\hline
Percent\_Storage\_dist & 0.00 & 0.00 & 0.01 & 0.26\\
\hline
Percent\_Factory\_dist & 0.00 & 0.00 & 0.00 & 0.04\\
\hline
Percent\_Other\_dist & 0.00 & 0.00 & 0.00 & 0.09\\
\hline
Percent\_Com\_basic\_mean & 0.00 & 0.04 & 0.07 & 0.54\\
\hline
Percent\_Res\_basic\_mean & 0.00 & 0.46 & 0.43 & 0.66\\
\hline
Percent\_Office\_basic\_mean & 0.00 & 0.01 & 0.03 & 0.44\\
\hline
Percent\_Retail\_basic\_mean & 0.00 & 0.02 & 0.02 & 0.08\\
\hline
Percent\_Garage\_basic\_mean & 0.00 & 0.00 & 0.00 & 0.29\\
\hline
Percent\_Storage\_basic\_mean & 0.00 & 0.00 & 0.01 & 0.23\\
\hline
Percent\_Factory\_basic\_mean & 0.00 & 0.00 & 0.00 & 0.03\\
\hline
Percent\_Other\_basic\_mean & 0.00 & 0.00 & 0.00 & 0.04\\
\hline
Percent\_Com\_dist\_perc\_change & -0.90 & 0.00 & 0.00 & 6.18\\
\hline
Percent\_Res\_dist\_perc\_change & -0.50 & 0.00 & 0.03 & 36.73\\
\hline
Percent\_Office\_dist\_perc\_change & -1.00 & 0.00 &  & \\
\hline
Percent\_Retail\_dist\_perc\_change & -0.82 & 0.00 &  & \\
\hline
Percent\_Garage\_dist\_perc\_change & -1.00 & 0.00 &  & \\
\hline
Percent\_Storage\_dist\_perc\_change & -1.00 & -0.01 &  & \\
\hline
Percent\_Factory\_dist\_perc\_change & -1.00 & 0.00 &  & \\
\hline
Percent\_Other\_dist\_perc\_change & -1.00 & 0.00 &  & \\
\hline
SMA\_Price\_2\_year\_dist & 0.00 & 400.01 & 496.30 & 3,816.57\\
\hline
SMA\_Price\_3\_year\_dist & 0.00 & 396.94 & 492.00 & 3,816.57\\
\hline
SMA\_Price\_5\_year\_dist & 8.83 & 425.55 & 515.29 & 3,877.53\\
\hline
Percent\_Change\_SMA\_2\_dist & -0.13 & 0.03 & 552.33 & 804,350.67\\
\hline
Percent\_Change\_SMA\_5\_dist & -0.09 & 0.02 & 317.46 & 322,504.58\\
\hline
EMA\_Price\_2\_year\_dist & 0.00 & 378.63 & 475.54 & 3,431.17\\
\hline
EMA\_Price\_3\_year\_dist & 8.83 & 382.25 & 476.05 & 3,296.46\\
\hline
EMA\_Price\_5\_year\_dist & 7.88 & 386.34 & 468.91 & 2,813.34\\
\hline
Percent\_Change\_EMA\_2\_dist & -0.09 & 0.06 & 346.51 & 480,829.57\\
\hline
Percent\_Change\_EMA\_5\_dist & -0.02 & 0.06 & 303.55 & 273,458.42\\
\hline
SMA\_Price\_2\_year\_basic\_mean & 0.02 & 412.46 & 496.75 & 2,509.79\\
\hline
SMA\_Price\_3\_year\_basic\_mean & 0.02 & 409.00 & 492.43 & 2,509.79\\
\hline
SMA\_Price\_5\_year\_basic\_mean & 17.16 & 443.34 & 515.67 & 2,621.01\\
\hline
Percent\_Change\_SMA\_2\_basic\_mean & -0.13 & 0.04 & 543.51 & 393,749.99\\
\hline
Percent\_Change\_SMA\_5\_basic\_mean & -0.09 & 0.03 & 312.46 & 157,500.00\\
\hline
EMA\_Price\_2\_year\_basic\_mean & 0.02 & 390.30 & 475.96 & 2,259.21\\
\hline
EMA\_Price\_3\_year\_basic\_mean & 11.39 & 393.25 & 476.45 & 2,136.36\\
\hline
EMA\_Price\_5\_year\_basic\_mean & 15.30 & 402.06 & 469.09 & 1,848.27\\
\hline
Percent\_Change\_EMA\_2\_basic\_mean & -0.09 & 0.06 & 340.89 & 235,378.24\\
\hline
Percent\_Change\_EMA\_5\_basic\_mean & -0.02 & 0.06 & 296.78 & 133,547.59\\
\hline
SMA\_Price\_2\_year\_dist\_perc\_change & -0.74 & 0.05 & 0.17 & 10,540.56\\
\hline
SMA\_Price\_3\_year\_dist\_perc\_change & -0.74 & 0.05 & 0.17 & 10,540.56\\
\hline
SMA\_Price\_5\_year\_dist\_perc\_change & -0.74 & 0.04 & 0.06 & 15.37\\
\hline
Percent\_Change\_SMA\_2\_dist\_perc\_change & -Inf & -0.24 & NaN & \\
\hline
Percent\_Change\_SMA\_5\_dist\_perc\_change & -Inf & -0.14 & NaN & \\
\hline
EMA\_Price\_2\_year\_dist\_perc\_change & -0.74 & 0.06 & 0.18 & 10,540.57\\
\hline
EMA\_Price\_3\_year\_dist\_perc\_change & -0.73 & 0.06 & 0.08 & 15.06\\
\hline
EMA\_Price\_5\_year\_dist\_perc\_change & -0.63 & 0.06 & 0.07 & 12.04\\
\hline
Percent\_Change\_EMA\_2\_dist\_perc\_change & -Inf & -0.13 & NaN & \\
\hline
Percent\_Change\_EMA\_5\_dist\_perc\_change & -556.60 & -0.10 &  & \\
\hline
SMA\_Price\_2\_year\_basic\_mean\_perc\_change & -0.55 & 0.05 & 0.12 & 9,375.77\\
\hline
SMA\_Price\_3\_year\_basic\_mean\_perc\_change & -0.55 & 0.05 & 0.11 & 9,375.77\\
\hline
SMA\_Price\_5\_year\_basic\_mean\_perc\_change & -0.50 & 0.04 & 0.06 & 5.90\\
\hline
Percent\_Change\_SMA\_2\_basic\_mean\_perc\_change & -Inf & -0.19 & NaN & \\
\hline
Percent\_Change\_SMA\_5\_basic\_mean\_perc\_change & -Inf & -0.12 & NaN & \\
\hline
EMA\_Price\_2\_year\_basic\_mean\_perc\_change & -0.53 & 0.06 & 0.12 & 9,375.78\\
\hline
EMA\_Price\_3\_year\_basic\_mean\_perc\_change & -0.47 & 0.06 & 0.08 & 23.54\\
\hline
EMA\_Price\_5\_year\_basic\_mean\_perc\_change & -0.37 & 0.06 & 0.07 & 4.81\\
\hline
Percent\_Change\_EMA\_2\_basic\_mean\_perc\_change & -Inf & -0.13 & NaN & \\
\hline
Percent\_Change\_EMA\_5\_basic\_mean\_perc\_change & -136.59 & -0.11 &  & \\
\hline
\end{tabular}}
\end{table}

\hypertarget{dependent-variables}{%
\subsection{Dependent Variables}\label{dependent-variables}}

The final step in creating the modeling data was to define the dependent
variables reflective of the prediction tasks; a binary variable for
classification and a continuous variable for regression:

\begin{enumerate}
\def\labelenumi{\arabic{enumi})}
\tightlist
\item
  \textbf{Binary: Sold} whether a tax lot sold in a given year. Used in
  the Probability of Sale classification model.
\item
  \textbf{Continuous: Sale-Price-per-SF} The price-per-square-foot
  associated with a transaction, if a sale took place. Used in the Sale
  Price Regression model.
\end{enumerate}

\noindent Table \ref{tab:OutcomeDistro} describes the distributions of
both outcome variables.

\begin{table}[t]

\caption{\label{tab:table 4}\label{tab:OutcomeDistro} Distributions for Outcome Variables}
\centering
\resizebox{\linewidth}{!}{
\begin{tabular}{l|l|l}
\hline
  & Sold & Sale Price per SF\\
\hline
Min. & 0.00 & 0.0\\
\hline
1st Qu. & 0.00 & 163.5\\
\hline
Median & 0.00 & 375.2\\
\hline
Mean & 0.04 & 644.8\\
\hline
3rd Qu. & 0.00 & 783.3\\
\hline
Max. & 1.00 & 83,598.7\\
\hline
\end{tabular}}
\end{table}

\hypertarget{algorithms-comparison}{%
\subsection{Algorithms Comparison}\label{algorithms-comparison}}

We implemented and compared several algorithms across our two-stage
process. In Stage 1, the Random Forest algorithm was used to identify
the optimal subset of building types and geographies for our spatial lag
aggregation assumptions. In Stage 2, we analyzed the hold-out test
performance of several algorithms including Random Forest, generalized
linear model (GLM), gradient boosting machine (GBM), and feed-forward
artificial neural network (ANN). Each algorithm was run over the three
competing feature engineering datasets and for both the classification
and regression tasks.

\hypertarget{random-forest}{%
\subsubsection{Random Forest}\label{random-forest}}

Random Forest was proposed by Breiman (2001) as an ensemble of
prediction decision trees iteratively trained across randomly generated
subsets of data. Algorithm \ref{alg:RandomForestAlo} outlines the
procedure (Hastie, Tibshirani, \& Friedman, 2001).

\begin{algorithm}
\caption{Random Forest for Regression or Classification}\label{alg:RandomForestAlo}

\begin{enumerate}
\item For $b = 1$ to $B$
\begin{enumerate}
\item Draw a bootstrap sample $Z$ of the size $N$ from the training data.
\item Grow a random-forest tree $T_b$ to the bootstrapped data, by recursively repeating the following steps for each terminal node of the tree, until the minimum node size $n_{min}$ is reached.
\begin{enumerate}
\item Select $m$ variables at random from the $p$ variables
\item Pick the best variable/split-point among the $m$.
\item Split the node into two daughter nodes.
\end{enumerate}
\end{enumerate}
\item Output the ensemble of trees $\{T_b\}_1^B$.
\end{enumerate}

To make a prediction at a new point $x$:

\textit{Regression:} $\hat{f}_{rf}^B(x) = \frac{1}{B}\sum_{b=1}^{B}T_b(x)$

\textit{Classification:} Let $\hat{C_b}(x)$ be the class prediction of the $b$th random-forest tree. Then $\hat{C_{rf}^B}(x)=$ \textit{majority vote} $\{\hat{C}_{b}(x)\}_1^B $

\end{algorithm}

Previous works have found the Random Forest algorithm suitable to
prediction tasks involving real estate (Antipov \& Pokryshevskaya, 2012;
Schernthanner et al., 2016). While algorithms exist that may outperform
Random Forest in terms of predictive accuracy (such as neural networks
and functional gradient descent algorithms), Random Forest is highly
scalable and parallelizable, and is, therefore, an attractive choice for
quickly assessing the predictive power of different feature engineering
techniques. For these reasons and more outlined below, we selected
Random Forest as the algorithm for Stage 1 of our modeling process.

Random Forest, like all predictive algorithms used in this work, suits
both classification and regression tasks. The Random Forest algorithm
works by generating a large number of independent classification or
regression decision trees and then employing majority voting (for
classification) or averaging (for regression) to generate predictions.
Over a dataset of N rows by M predictors, a bootstrap sample of the data
is chosen (n \textless{} N) as well as a subset of the predictors (m
\textless{} M). Individual decision or regression trees are built on the
n by m sample. Because the trees develop independently (and not
sequentially, as is the case with most functional gradient descent
algorithms), the tree building process can be executed in parallel. With
a sufficiently large number of computer cores, the model training time
can be significantly reduced.

We chose Random Forest as the algorithm for Stage 1 because:

\begin{enumerate}
\def\labelenumi{\arabic{enumi})}
\tightlist
\item
  The algorithm can be parallelized and is relatively fast compared to
  neural networks and functional gradient descent algorithms
\item
  Can accommodate categorical variables with many levels. Real estate
  data often contains information describing the location of the
  property, or the property itself, as one of a large set of possible
  choices, such as neighborhood, county, census tract, district,
  property type, and zoning information. Because factors need to be
  recoded as individual dummy variables in the model building process,
  factors with many levels quickly encounter the curse of dimensionality
  in multiple regression techniques.
\item
  Appropriately handles missing data. Predictions can be made with the
  parts of the tree which are successfully built, and therefore, there
  is no need to filter out incomplete observations or impute missing
  values. Since much real estate data is self-reported, incomplete
  fields are common in the data.
\item
  Robust against outliers. Because of bootstrap sampling, outliers
  appear in individual trees less often, and therefore, their influence
  is curtailed. Real estate data, especially with regards to pricing,
  tends to contain outliers. For example, the dependent variable in one
  of our models, sale price, shows a clear divergence in the median and
  mean, as well as a maximum significantly higher than the third
  quartile.
\item
  Can recognize non-linear relationships in data, which is useful when
  modeling spatial relationships.
\item
  Is not affected by co-linearity in the data. This is highly valuable
  as real estate data can be highly correlated.
\end{enumerate}

To run the model, we chose the h2o.randomForest implementation from the
h2o R open source library. The h2o implementation of the Random Forest
algorithm is particularly well-suited for high parallelization. For more
information, see \url{https://www.h2o.ai/}.

\hypertarget{generalized-linear-model}{%
\subsubsection{Generalized Linear
Model}\label{generalized-linear-model}}

A generalized linear model (GLM) is an extension of the general linear
model that estimates an independent variable \(y\) as the linear
combination of one or more predictor variables. The dependent variable
\(y\) for observation \(i\) \((i = 1, 2, ..., n)\) is modeled as a
linear function of \((p - 1)\) independent variables
\(x1, x2,... ,xp-1\) as

\[
y_i = \beta_0+\beta_1x_{i1}+...+\beta_{p-1}x_{i(p-1)}+e_i
\]

A GLM is composed of three primary parts: a linear model, a link
function and a variance function. The linear model takes the form
\(\eta_i = \beta_0+\beta_1x_{i1}+...+\beta_{p}x_{ip}\). The link
function, \(g(\mu)=\eta\) relates the mean to the linear model, and the
variance function \(Var(Y) = \phi V(\mu)\) relates the model variance to
the mean (Hoffmann, 2004; Turner, 2008).

Several family types of GLM's exist. For a binary independent variable,
a binomial logistic regression is appropriate. For a continuous
independent variable, the Gaussian or another distribution is
appropriate. For our purposes, the Gaussian family is used for our
regression task and binomial for the classification.

\hypertarget{gradient-boosting-machine}{%
\subsubsection{Gradient Boosting
Machine}\label{gradient-boosting-machine}}

Gradient boosting machine (GBM) is one of the most popular machine
learning algorithms available today. The algorithm uses iteratively
refined approximations, obtained through cross-validation, to
incrementally increase predictive accuracy. Similar to Random Forest,
GBM is an ensemble technique that builds and averages many regression
models together. Unlike Random Forest, GBM incrementally improves each
successive iteration by following the gradient of the loss function at
each step (Friedman, 1999). The algorithm we used, which is the
tree-variant of the generic gradient boosting algorithm, is outlined in
algorithm \ref{alg:GBMAlo} (Hastie et al., 2001 pg. 361).

\begin{algorithm}
\caption{Gradient Tree Boosting Algorithm}\label{alg:GBMAlo}
\begin{enumerate}
\item Initialize: $f_0(x) = \arg\min_\gamma \sum_{i=1}^N L(y_i, \gamma).$
\item For $m$ = 1 to $M$:
\begin{enumerate}
\item For $1,2,...,N$ compute "pseudo-residuals":
$$
r_{im} = -\left[\frac{\partial L(y_i, f(x_i))}{\partial f(x_i)}\right]_{f=f_{m-1}(x)} \quad
$$
\item Fit a regression tree to the targets $r_{im}$ giving terminal regions $R_{jm}, j = 1,2,...J_m$
\item For $j = 1,2,...,J_m$ compute:

$$
\gamma_{jm} = \underset{\gamma}{\operatorname{arg\,min}} \sum_{xi \in R_{jm}} L\left(y_i, f_{m-1}(x_i) + \gamma \right).
$$
\item Update $f_m(x) = F_{m-1}(x) + \sum_{j=1}^{J_m}\gamma_{jm}I(x \in R_{jm})$
\end{enumerate}
\item Output $\hat{f}(x) = f_m(x)$
\end{enumerate}
\end{algorithm}

\hypertarget{feed-forward-artificial-neural-network}{%
\subsubsection{Feed-Forward Artificial Neural
Network}\label{feed-forward-artificial-neural-network}}

The artificial neural network (ANN) implementation used in this work is
a multi-layer feed-forward artificial neural network. Common synonyms
for ANN models are multi-layer perceptrons and, more recently, deep
neural networks. The feed-forward ANN is one of the most common neural
network algorithms, but other types exist, such as the convolutional
neural network (CNN) which performs well on image classification tasks,
and the recurrent neural network (RNN) which is well-suited for
sequential data such as text and audio (Schmidhuber, 2015). The
feed-forward ANN is typically best suited for tabular data.

A neural network model is made up of an input layer made up of raw data,
one or more hidden layers used for transformations, and an output layer.
At each hidden layer, the input variables are combined using varying
weights with all other input variables. The output from one hidden layer
is then used as the input to the next layer, and so on. Tuning a neural
network is the process of refining the weights to minimize a loss
function and make the model fit the training data well (Hastie et al.,
2001).

For both our classification and regression tasks, we use sum-of-squared
errors as our error function, and we tune the set of weights \(\theta\)
to minimize:

\[
R(\theta)=\sum_{k=1}^K\sum_{i=1}^N(y_{ik}-f_{k}(x_i))^2
\] A typical approach to minimizing \(R(\theta)\) is by gradient
descent, called back-propagation in this setting (Hastie et al., 2001).
The algorithm iteratively tunes weighting values back and forth across
the hidden layers in accordance with the gradient descent of the loss
function until material improvement can no longer happen or the
algorithm reaches a user-defined limit.

For our implementation, we used the rectifier activation function with
1024 hidden layers, 100 epochs and L1 regularization set to 0.00001. The
implementation we chose was the h2o.deeplearning open source R library.
For more information, see \url{https://www.h2o.ai/}.

\hypertarget{model-validation}{%
\subsection{Model Validation}\label{model-validation}}

\begin{figure}
\centering
\includegraphics{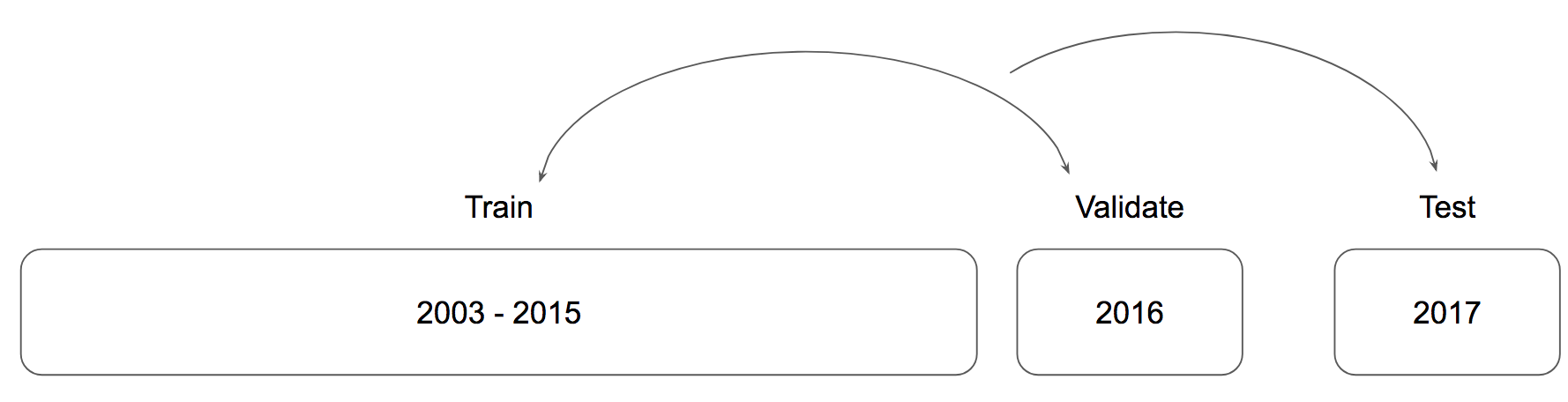}
\caption{\label{fig:Train Test Validate}Spatial Out-of-time validation}
\end{figure}

Our goal was to be able to successfully predict both the probability and
amount of real estate sales into the near future. As such, we trained
and evaluated our models using out-of-time validation to assess
performance. As shown in Figure \ref{fig:Train Test Validate} The models
were trained using data from 2003-2015. We used 2016 data during the
training process for model validation purposes. Finally, we scored our
models using 2017 data as a hold-out sample. Using out-of-time
validation ensured that our models generalized well into the immediate
future.

\hypertarget{evaluation-metrics}{%
\subsection{Evaluation Metrics}\label{evaluation-metrics}}

We chose evaluation metrics that allowed us to easily compare the
performance of the models against other similar models with the same
dependent variable. The classification models (probability of sale) were
compared using the area under the ROC curve (AUC). The regression models
(sale price) were compared using root mean squared errors (RMSE). Both
evaluation metrics are common for their respective outcome variable
types, and as such were useful for comparing within model-groups.

\hypertarget{area-under-the-roc-curve}{%
\subsubsection{Area Under the ROC
Curve}\label{area-under-the-roc-curve}}

A classification model typically outputs a probability that a given case
in the data belongs to a group. In the case of binary classification,
the value falls between 0 and 1. There are many techniques for
determining the cut off threshold for classification; a typical method
is to assign anything above a 0.5 into the 1 or positive class. An ROC
curve (receiver operating characteristic curve) plots the True Positive
Rate vs.~the False Positive rate at different classification thresholds;
it is a measurement of the performance of a classification model across
all possible thresholds and therefore sidesteps the need to assign a
cutoff arbitrarily.

AUC is the integration of the ROC curve from (0,0) to (1,1), or
\(AUC = \int_{(0,0)}^{(1,1)} f(x)dx\). A value of 0.5 represents a
perfectly random model, while a value of 1.0 represents a model that can
perfectly discriminate between the two classes. AUC is useful for
comparing classification models against one another because they are
both scale and threshold-invariant.

One of the drawbacks to AUC is that it does not describe the trade-offs
between false positives and false negatives. In certain circumstances, a
false positive might be considerably less desirable than a false
negative, or vice-versa. For our purposes, we rank false positives and
false negatives as equally undesirable outcomes.

\hypertarget{root-mean-squared-error}{%
\subsubsection{Root Mean Squared Error}\label{root-mean-squared-error}}

RMSE is a common measurement of the differences between regression model
predicted values and observed values. It is formally defined as
\(RMSE = \sqrt{ \frac{\sum_{1}^{T} (\hat{y}_t - y_t)^2}{T} }\), where
\(\hat{y}\) represents the prediction and \(y\) represents the observed
value at observation \(t\).

Lower RMSE scores are typically more desirable. An RMSE value of 0 would
indicate a perfect fit to the data. RMSE can be difficult to interpret
on its own; however, it is useful for comparing models with similar
outcome variables. In our case, the outcome variables
(sale-price-per-square-foot) are consistent across modeling datasets,
and therefore can be reasonably compared using RMSE.

\hypertarget{results}{%
\section{Results}\label{results}}

\hypertarget{summary-of-results}{%
\subsection{Summary of Results}\label{summary-of-results}}

We have conducted comparative analyses across a two-stage modeling
process. In Stage 1, using the Random Forest algorithm, we tested 3
competing feature engineering techniques (base, zip code aggregation,
and spatial lag aggregation) for both a classification task (predicting
the occurrence of a building sale) and a regression task (predicting the
sale price of a building). We analyzed the results of the first stage to
identify which geographies and building types our model assumptions
worked best. In Stage 2, using a subset of the modeling data (selected
via an analysis of the output from Stage 1), we compared four algorithms
-- GLM, Random Forest, GBM and ANN -- across our 3 competing feature
engineering techniques for both classification and regression tasks. We
analyzed the performance of the different model/data combos as well as
conducted an analysis of the variable importances for the top performing
models.

In Stage 1 (Random Forest, using all data), we found that models which
utilized spatial features outperformed those models using zip code
features the majority of the time for both classification and
regression. Of three models, the sale price regression model using
spatial features finished 1st or 2nd 24.1\% of the time (using RMSE as a
ranking criterion), while the zip code regression model finished in the
top two spots only 11.2\% of the time. Both models performed worse than
the base regression model overall, which ranked in 1st or 2nd place
31.5\% of the time. The story for the classification models was largely
the same: the spatial features tended to outperform the zip code data
while the base data won out overall. All models had similar performances
on training data, but the spatial and zip code datasets tended to
underperform when generalizing to the hold-out test data, suggesting
problems with overfitting.

We then analyzed the performance of both the regression and
classification Random Forest models by geography and building type. We
found that the models performed considerably better on walk up
apartments and elevator buildings (building types C and D) and in
Manhattan, Brooklyn and the Bronx. Using these as filtering criteria, we
created a subset of the modeling data for the subsequent modeling stage.

During Stage 2 (many algorithms using a subset of modeling data), we
compared four algorithms across the same three competing feature
engineering techniques using a filtered subset of the original modeling
data. Unequivocally, the spatial features performed best across all
models and tasks. For the classification task, the GBM algorithms
performed best in terms of AUC, followed by ANN and Random Forest. For
regression, the ANN algorithms performed best (as measured by RMSE as
well as Mean Absolute Error and R-squared) with the spatial features ANN
model performing best.

We conclude that spatial lag features can significantly increase the
accuracy of machine learning-based real estate sale prediction models.
We find that model overfitting presents a challenge when using spatial
features, but that this can be overcome by implementing different
algorithms, specifically ANN and GBM. Finally, we find that our
implementation of spatial lag features works best for certain kinds of
buildings in specific geographic areas, and we hypothesize that this is
due to the assumptions made when building the spatial features.

\hypertarget{stage-1-random-forest-models-using-all-data}{%
\subsection{Stage 1) Random Forest Models Using All
Data}\label{stage-1-random-forest-models-using-all-data}}

\hypertarget{sale-price-regression-models}{%
\subsubsection{Sale Price Regression
Models}\label{sale-price-regression-models}}

We analyzed the RMSE of the Random Forest models predicting sale price
across feature engineering methods, borough and building type. Table
\ref{tab:SalePriceModelRank} displays the average ranking by model type
as well as the distribution of models that ranked first, second and
third for each respective borough/building type combination. When we
rank the models by performance for each borough, building type
combination, we find that the spatial lag models outperform the zip code
models in 72\% of cases with an average model-rank of 2.11 and 2.5,
respectively.

\begin{table}[t]

\caption{\label{tab:Sale Price Model Rank Distributions}\label{tab:SalePriceModelRank} Sale Price Model Rankings, RMSE by Borough and Building Type}
\centering
\begin{tabular}{l|l|l|l|r}
\hline
Model Rank & 1 & 2 & 3 & Average Rank\\
\hline
Base & 22.2\% & 9.3\% & 1.9\% & 1.39\\
\hline
Spatial Lag & 5.6\% & 18.5\% & 9.3\% & 2.11\\
\hline
Zip & 5.6\% & 5.6\% & 22.2\% & 2.50\\
\hline
\end{tabular}
\end{table}

\begin{figure}
\centering
\includegraphics{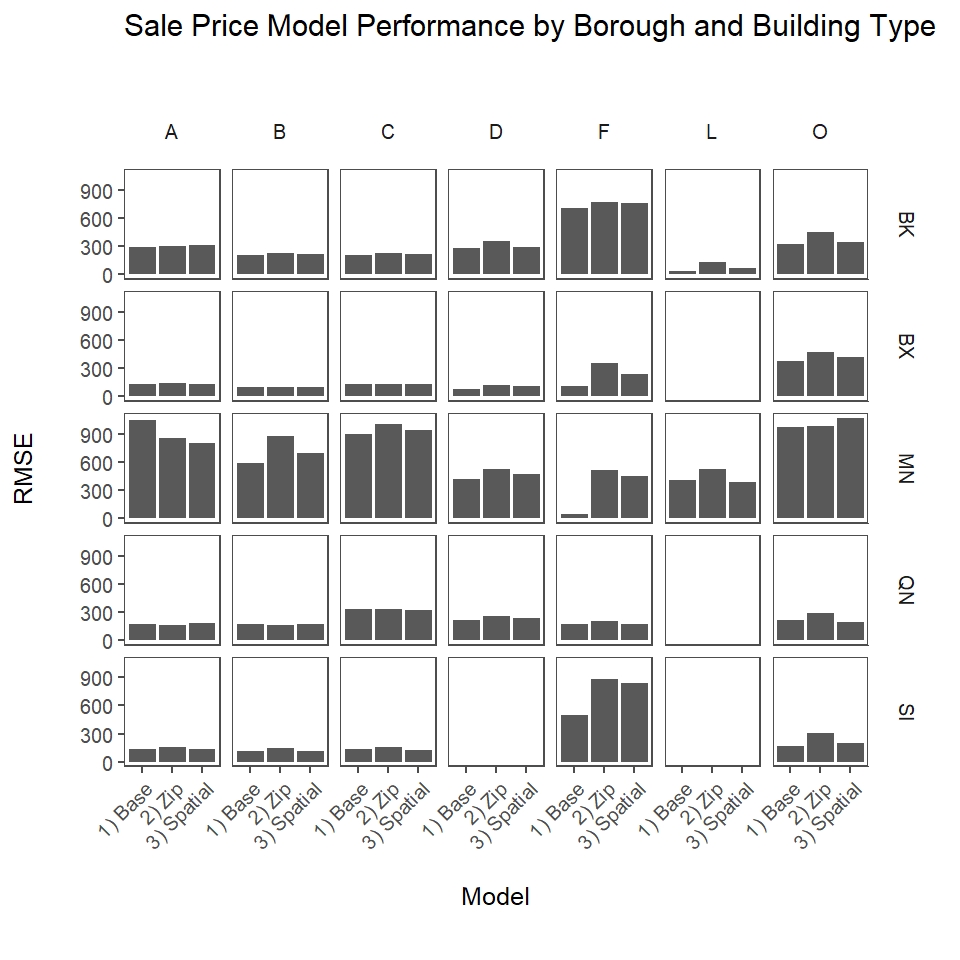}
\caption{\label{fig:RMSE by boro and build type}RMSE By Borough and
Building Type}
\end{figure}

The base modeling dataset tends to outperform both enriched datasets,
suggesting an issue with model overfitting in some areas. We see further
evidence of overfitting in Table \ref{tab:SalePriceEval} where, despite
similar performances on the validation data, the zip and spatial models
have higher validation-to-test-set spreads. Despite this, the spatial
lag features outperform all other models in specific locations, notably
in Manhattan as shown in Figure \ref{fig:RMSE by boro and build type}.

\begin{table}[t]

\caption{\label{tab:Sale Price Evaluations}\label{tab:SalePriceEval} Sale Price Model RMSE For Validation and Test Hold-out Data}
\centering
\begin{tabular}{l|r|r|r}
\hline
type & base & zip & spatial lag\\
\hline
Validation & 280.63 & 297.97 & 286.23\\
\hline
Test & 287.83 & 300.60 & 297.92\\
\hline
\end{tabular}
\end{table}

Figure \ref{fig:RMSE by boro and build type} displays test RMSE by
model, faceted by borough on the y-axis and building type on the x-axis
(See Table \ref{tab:categoryTable} and Table \ref{tab:by_class} for a
description of building type codes). We make the following observations
from Figure \ref{fig:RMSE by boro and build type}:

\begin{itemize}
\tightlist
\item
  The spatial modeling data outperforms both base and zip code in 6
  cases, notably for type A buildings (one family dwellings) and type L
  buildings (lofts) in Manhattan as well as type O buildings (offices)
  in Queens
\item
  The ``residential'' building types A (one-family dwellings), B
  (two-family dwellings), C (walk up apartments) and D (elevator
  apartments) have lower RMSE scores compared to the non-residential
  types
\item
  Spatial features perform best in Brooklyn, the Bronx, and Manhattan
  and for residential building types
\end{itemize}

\hypertarget{probability-of-sale-classification-models}{%
\subsubsection{Probability of Sale Classification
Models}\label{probability-of-sale-classification-models}}

Similar to the results of the sale price regression models, we found
that the spatial models performed better on the hold-out test data
compared to the zip code data, as shown in Table
\ref{tab:ProbSaleModelAUC}. The base modeling data continued to
outperform the spatial and zip code data overall.

\begin{table}[t]

\caption{\label{tab:Prob Model AUC}\label{tab:ProbSaleModelAUC} Probability of Sale Model AUC}
\centering
\begin{tabular}{l|r|r|r}
\hline
Model AUC & Base & Zip & Spatial Lag\\
\hline
Validation & 0.832 & 0.829 & 0.829\\
\hline
Test & 0.830 & 0.825 & 0.828\\
\hline
\end{tabular}
\end{table}

\begin{figure}
\centering
\includegraphics{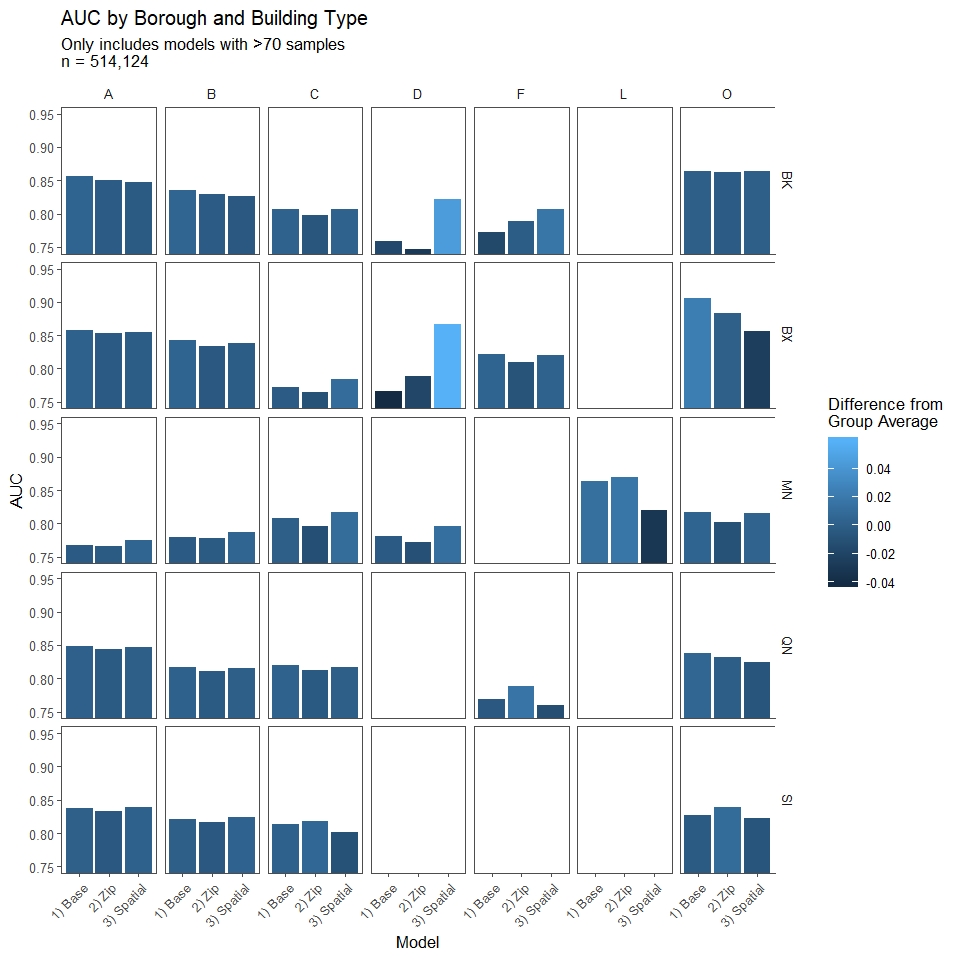}
\caption{\label{fig:AUC by boro and build type}AUC By Borough and
Building Type}
\end{figure}

Figure \ref{fig:AUC by boro and build type} shows a breakdown of model
AUC faceted along the x-axis by building type and along the y-axis by
borough. The coloring indicates by how much a model's AUC diverges from
the cell average, which is useful for spotting over performers. We
observed the following from Figure \ref{fig:AUC by boro and build type}:

\begin{itemize}
\tightlist
\item
  The spatial models outperform all other models for elevator buildings
  (type D) and walk up apartments (type C), particularly in Brooklyn,
  the Bronx, and Manhattan
\item
  Classification tends to perform poorly in Manhattan vs.~other Boroughs
\item
  The spatial models perform well in Manhattan for the residential
  building types (A, B, C, and D)
\end{itemize}

If we rank the classification models' performance for each borough and
building type, we see that the spatial models consistently outperform
the zip code models, as shown in Table \ref{tab:ProbModelAUCRank}. From
this (as well as from similar patterns seen in the regression models) we
infer that the spatial data is a superior data engineering technique;
however, the algorithm used needs to account for potential model
overfitting. In the next section, we discuss refining the data used as
well as employing different algorithms to maximize the predictive
capability of the spatial features.

\begin{table}[t]

\caption{\label{tab:Prob Model AUC Average Rank}\label{tab:ProbModelAUCRank} Distribution and Average Model Rank for Probability of Sale by AUC across Borough and Building Types}
\centering
\begin{tabular}{l|l|l|l|r}
\hline
Model Rank & 1 & 2 & 3 & Average Rank\\
\hline
Base & 16.2\% & 12.0\% & 5.1\% & 2.22\\
\hline
Spatial Lag & 11.1\% & 13.7\% & 8.5\% & 2.09\\
\hline
Zip & 6.0\% & 7.7\% & 19.7\% & 1.69\\
\hline
\end{tabular}
\end{table}

\hypertarget{stage-2-model-comparisons-using-specific-geographies-and-building-types}{%
\subsection{Stage 2) Model Comparisons Using Specific Geographies and
Building
Types}\label{stage-2-model-comparisons-using-specific-geographies-and-building-types}}

Using the results from the first modeling exercise, we conclude that
walk up apartments and elevator buildings in Manhattan, Brooklyn and the
Bronx are suitable candidates for prediction using our current
assumptions. These buildings share the characteristics of being
residential as well as being reasonably uniform in their geographic
density. We analyze the performance of four algorithms (GLM, Random
Forest, GBM, and ANN), using three feature engineering techniques, for
both classification and regression, making the total number
\texttt{4\ x\ 3\ x\ 2\ =\ 24} models.

\hypertarget{regression-model-comparisons}{%
\subsubsection{Regression Model
Comparisons}\label{regression-model-comparisons}}

\begin{figure}
\centering
\includegraphics{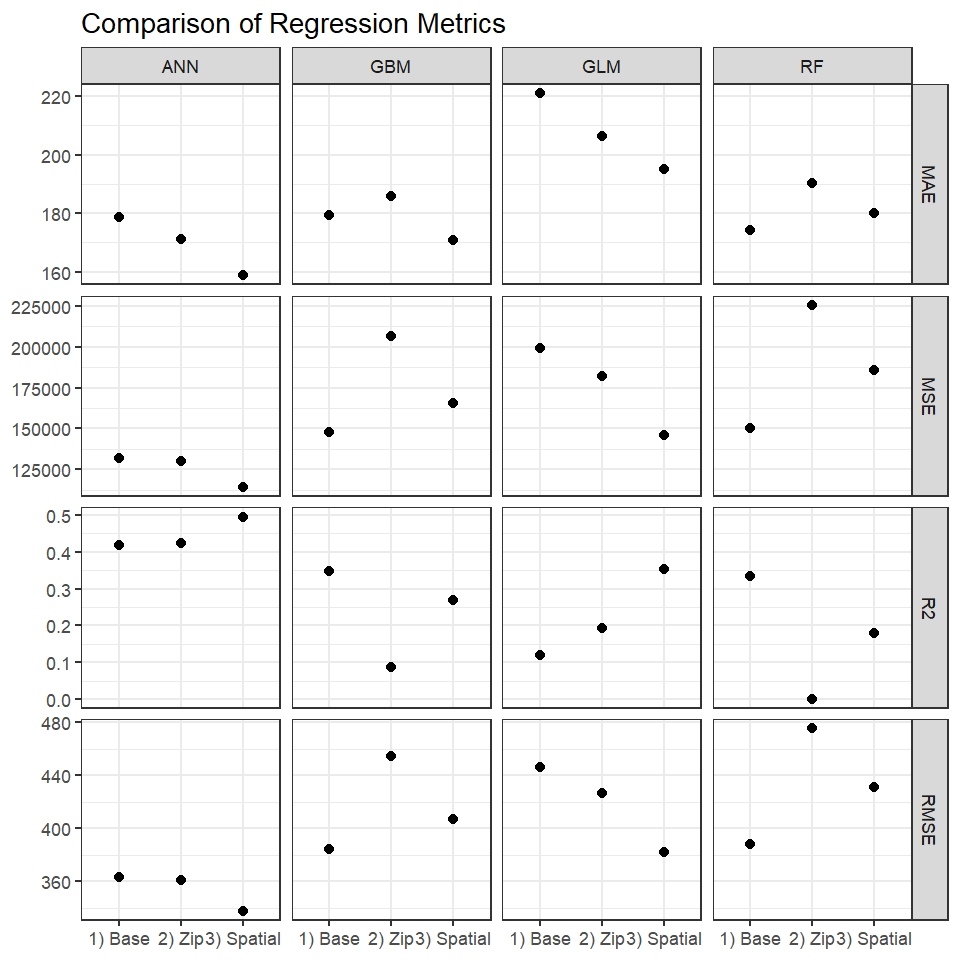}
\caption{\label{fig:Model RMSE Comparrison}Comparative Regression
Metrics}
\end{figure}

The predictive accuracies of the various regression models were
evaluated using RMSE, described in detail in the methodology section, as
well as Mean Absolute Error (MAE), Mean Squared Error (MSE) and
R-Squared. These four indicators were calculated using the hold-out test
data, which ensured that the models performed well when predicting sale
prices into the near future. The comparison metrics are presented in
Table \ref{tab:RegModelTable} and Figure
\ref{fig:Model RMSE Comparrison}. We make the following observations
about Table \ref{tab:RegModelTable} and Figure
\ref{fig:Model RMSE Comparrison}:

\begin{enumerate}
\def\labelenumi{\arabic{enumi})}
\tightlist
\item
  The ANN models perform best in nearly every metric across nearly all
  feature sets, with GBM a close second in some circumstances
\item
  ANN and GLM improve linearly in all metrics as you move from base to
  zip to spatial, with spatial performing the best. GBM and Random
  Forest, on the other hand, perform best on the base and spatial
  feature sets and poorly on the zip features
\item
  We see a similar pattern in the Random Forest results compared to the
  previous modeling exercise using the full dataset: the base features
  outperform both spatial and zip, with spatial coming in second
  consistently. This pattern further validates our reasoning that
  spatial features are highly predictive but suffer from overfitting and
  other algorithm-related reasons
\item
  The highest model R-squared is the ANN using spatial features at
  0.494, indicating that this model can account for nearly 50\% of the
  variance in the test data. Compared to the R-squared of the more
  traditional base GLM at 0.12, this represents a more than 3-fold
  improvement in predictive accuracy
\end{enumerate}

\begin{table}[t]

\caption{\label{tab:Reg Model RMSE Compare}\label{tab:RegModelTable} Prediction Accuracy of Regression Models on Test Data}
\centering
\begin{tabular}{l|l|r|r|r|r}
\hline
Data & Model & RMSE & MAE & MSE & R2\\
\hline
1) Base & GLM & 446.35 & 221.16 & 199227.6 & 0.12\\
\hline
2) Zip & GLM & 426.93 & 206.49 & 182270.1 & 0.19\\
\hline
3) Spatial & GLM & 382.32 & 195.00 & 146170.5 & 0.35\\
\hline
1) Base & RF & 387.99 & 174.24 & 150536.3 & 0.33\\
\hline
2) Zip & RF & 475.20 & 190.33 & 225811.7 & 0.00\\
\hline
3) Spatial & RF & 430.92 & 180.17 & 185695.5 & 0.18\\
\hline
1) Base & GBM & 384.11 & 179.27 & 147543.5 & 0.35\\
\hline
2) Zip & GBM & 454.53 & 186.00 & 206593.1 & 0.09\\
\hline
3) Spatial & GBM & 406.70 & 170.97 & 165408.0 & 0.27\\
\hline
1) Base & ANN & 363.02 & 178.58 & 131782.5 & 0.42\\
\hline
2) Zip & ANN & 360.88 & 171.22 & 130232.2 & 0.42\\
\hline
3) Spatial & ANN & 337.94 & 158.91 & 114202.0 & 0.49\\
\hline
\end{tabular}
\end{table}

\begin{figure}
\centering
\includegraphics{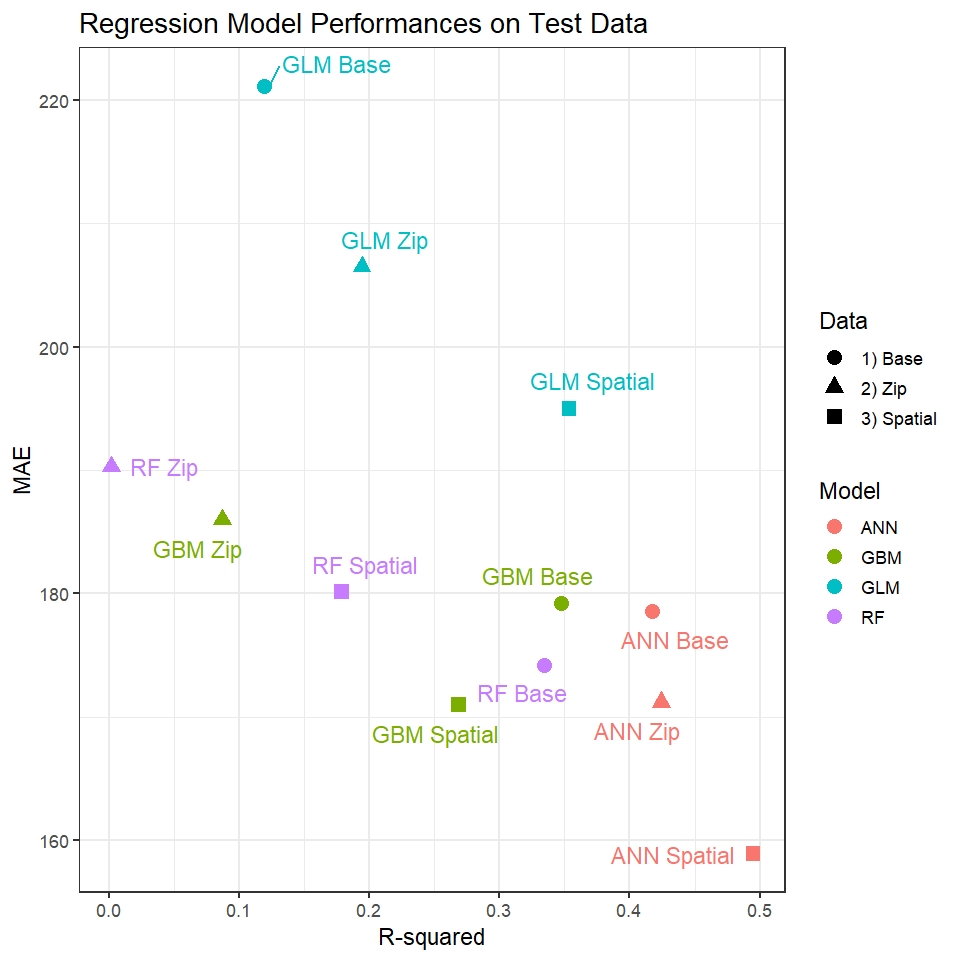}
\caption{\label{fig:reg Models Scatterplot}Regression Model Performances
On Test Data}
\end{figure}

\noindent Figure \ref{fig:reg Models Scatterplot} shows clusters of
model performances across R-squared and MAE, with the ANN models
outperforming their peers. This figure also makes clear that the
marriage of spatial features with the ANN algorithm results in a
dramatic reduction in error rate compared to the other techniques.

\hypertarget{classification-model-comparisons}{%
\subsubsection{Classification Model
Comparisons}\label{classification-model-comparisons}}

\begin{figure}
\centering
\includegraphics{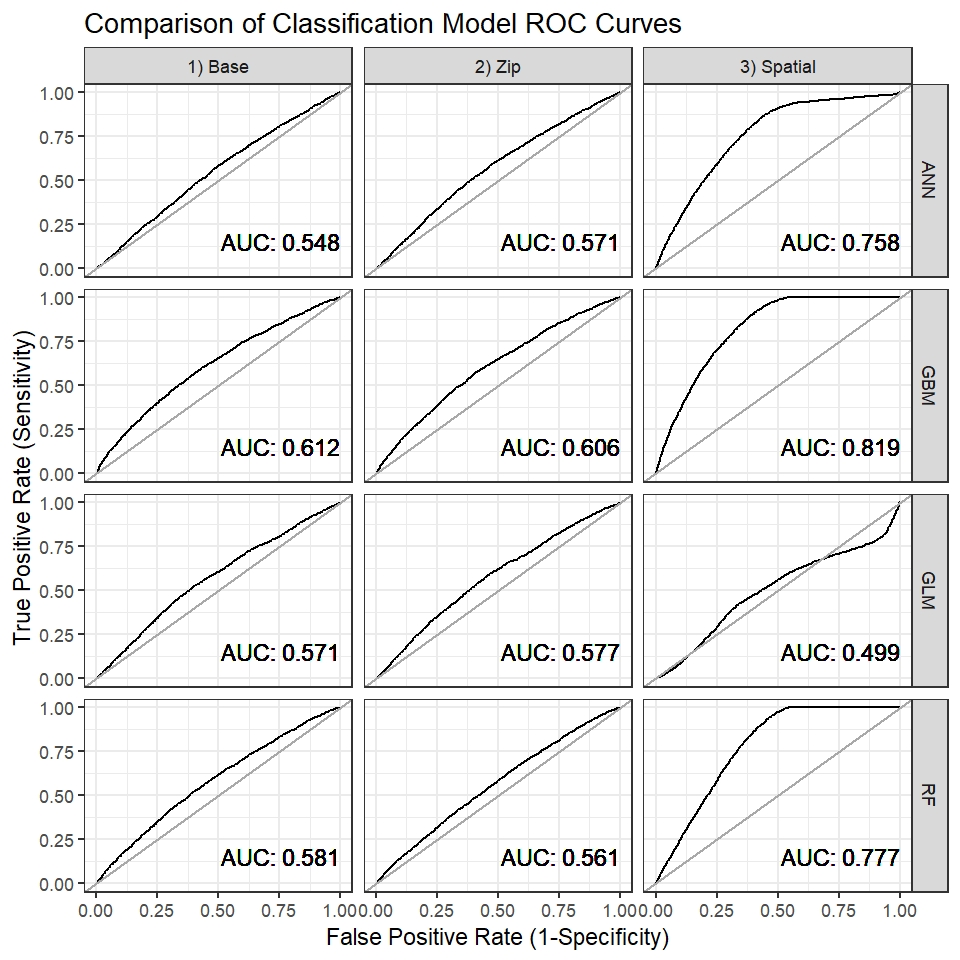}
\caption{\label{fig:Model AUC Comparrison}Comparison of Classification
Model ROC Curves}
\end{figure}

The classification models were assessed using AUC as well as MSE, RMSE,
and R-squared. As with the regression models, these four metrics were
calculated using the hold-out test data, ensuring that the models
generalize well into the near future. The comparison metrics are
presented in Table \ref{tab:ClassModelTable}. Figure
\ref{fig:Model AUC Comparrison} shows the ROC curves and corresponding
AUC for each algorithm/feature set combination.

\noindent We observe the following of Table \ref{tab:ClassModelTable}
and Figure \ref{fig:Model AUC Comparrison}:

\begin{enumerate}
\def\labelenumi{\arabic{enumi})}
\tightlist
\item
  Unlike the regression models, the GBM algorithm with spatial features
  proved to be the best performing classifier. All spatial models
  performed relatively well except the GLM spatial model
\item
  Only 3 models have positive R-squared values: ANN spatial, Random
  Forest spatial, and GLM base, indicating that these models are adept
  at predicting positive cases (occurrences of sales) in the test data
\item
  GLM spatial returned an AUC of less than 0.5, indicating a model that
  is conceptually worse than random. This is likely the result of
  overfitting
\end{enumerate}

\begin{table}[t]

\caption{\label{tab:Class Model Compare}\label{tab:ClassModelTable} Prediction Accuracy of Classification Models on Test Data}
\centering
\begin{tabular}{l|l|r|r|r|r}
\hline
Data & Model & AUC & MSE & RMSE & R2\\
\hline
1) Base & GLM & 0.57 & 0.03 & 0.17 & 0.00\\
\hline
2) Zip & GLM & 0.58 & 0.03 & 0.17 & 0.00\\
\hline
3) Spatial & GLM & 0.50 & 0.03 & 0.17 & -0.01\\
\hline
1) Base & RF & 0.58 & 0.03 & 0.17 & -0.03\\
\hline
2) Zip & RF & 0.56 & 0.03 & 0.17 & -0.06\\
\hline
3) Spatial & RF & 0.78 & 0.03 & 0.17 & 0.00\\
\hline
1) Base & GBM & 0.61 & 0.03 & 0.17 & -0.03\\
\hline
2) Zip & GBM & 0.61 & 0.03 & 0.17 & -0.03\\
\hline
3) Spatial & GBM & 0.82 & 0.03 & 0.16 & 0.04\\
\hline
1) Base & ANN & 0.55 & 0.03 & 0.17 & -0.03\\
\hline
2) Zip & ANN & 0.57 & 0.03 & 0.17 & -0.04\\
\hline
3) Spatial & ANN & 0.76 & 0.03 & 0.17 & -0.01\\
\hline
\end{tabular}
\end{table}

\begin{figure}
\centering
\includegraphics{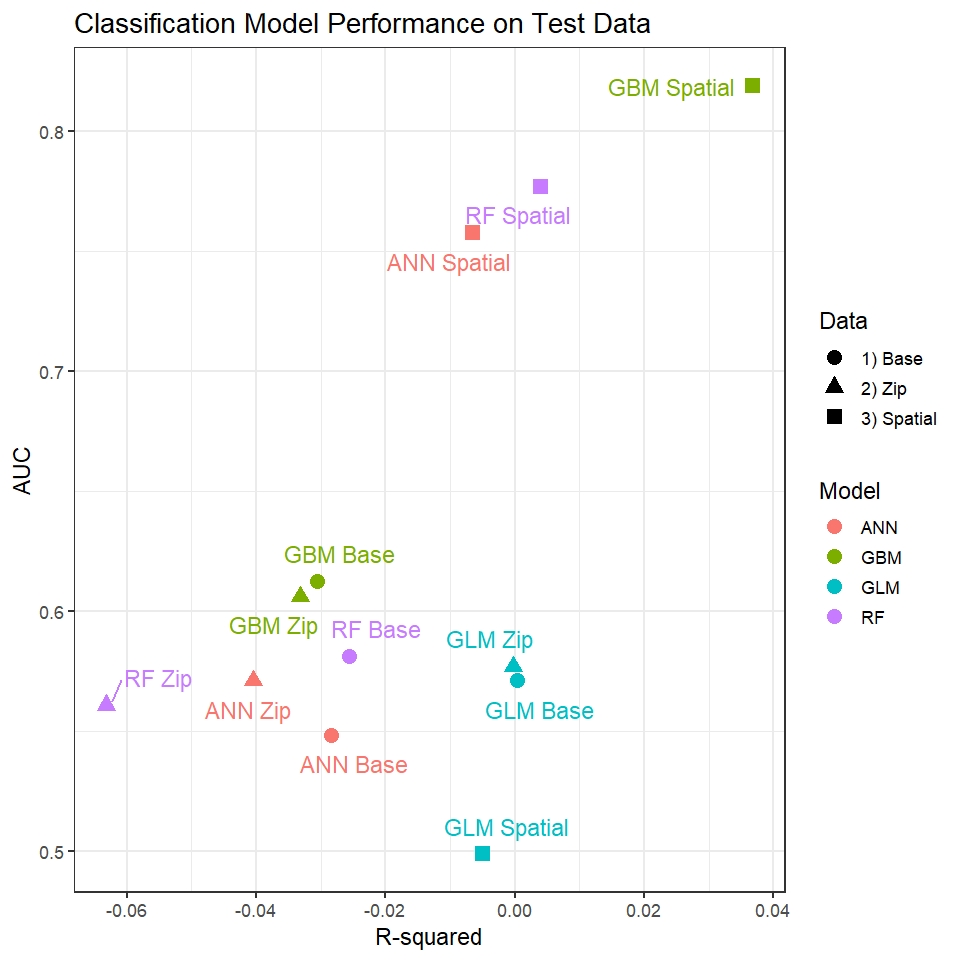}
\caption{\label{fig:Class Models Scatterplot}Scatterplot of
Classification Models}
\end{figure}

\noindent Figure \ref{fig:Class Models Scatterplot} plots the individual
models by AUC and R-squared. The spatial models tend to outperform the
other models by a significant margin. Interestingly, when compared to
the regression model scatterplot in Figure
\ref{fig:reg Models Scatterplot}, the classification models tend to
cluster by feature set. In \ref{fig:reg Models Scatterplot}, we see the
regression models clustering by algorithm.

\hypertarget{variable-importance-analysis-of-top-performing-models}{%
\subsection{Variable Importance Analysis of Top Performing
Models}\label{variable-importance-analysis-of-top-performing-models}}

We calculated the feature importance for each variable as the
proportional to the average decrease in the squared error after
including that variable in the model. The most important variable gets a
score of 1; scores for other variables are derived by standardizing
their measured reduction in error relative to the largest one. The top
10 variables for both the most successful regression and most successful
classification models are presented in Tables \ref{tab:RegVarImp} and
\ref{tab:ClassVarImp}.

\begin{table}[t]

\caption{\label{tab:Reg VarImp}\label{tab:RegVarImp} Feature Importance of Top Performing Regression Model}
\centering
\resizebox{\linewidth}{!}{
\begin{tabular}{l|l|r|l}
\hline
Variable & Description & Scaled Importance (Max = 1) & Cumulative \%\\
\hline
BuiltFAR & Floor area ratio built & 1.000 & 1.80\%\\
\hline
FacilFAR & Maximum Allowable Floor Area Ratio & 0.922 & 3.40\%\\
\hline
Last\_Sale\_Price\_Total & The previous sale price & 0.901 & 5.10\%\\
\hline
Last\_Sale\_Date & Date of last sale & 0.893 & 6.70\%\\
\hline
Last\_Sale\_Price & The previous sale price & 0.870 & 8.20\%\\
\hline
Years\_Since\_Last\_Sale & Number of years since last sale & 0.823 & 9.70\%\\
\hline
ResidFAR & Floor Area Ratio not yet built & 0.814 & 11.20\%\\
\hline
lon & Longitude & 0.773 & 12.60\%\\
\hline
Year & Year of record & 0.759 & 13.90\%\\
\hline
BldgDepth & Square feet from font to back & 0.758 & 15.30\%\\
\hline
\end{tabular}}
\end{table}

\begin{table}[t]

\caption{\label{tab:Class VarImp}\label{tab:ClassVarImp} Feature Importance of Top Performing Classification Model}
\centering
\resizebox{\linewidth}{!}{
\begin{tabular}{l|l|r|l}
\hline
Variable & Description & Scaled Importance (Max = 1) & Cumulative \%\\
\hline
Percent\_Neighbords\_Sold & Percent of Nearby Properties Sold in the Previous Year & 1.000 & 21.90\%\\
\hline
Percent\_Office & Percent of the build which is Office & 0.698 & 37.20\%\\
\hline
Percent\_Garage & Percent of the build which is Garage & 0.634 & 51.10\%\\
\hline
Percent\_Storage & Percent of the build which is Storage & 0.518 & 62.40\%\\
\hline
Building\_Age & The Age of the building & 0.225 & 67.40\%\\
\hline
Last\_Sale\_Price & Price of building last time is was sold & 0.165 & 71.00\%\\
\hline
Percent\_Retail & Percent of the build which is Retail & 0.147 & 74.20\%\\
\hline
Years\_Since\_Last\_Sale & Year since building last sold & 0.121 & 76.90\%\\
\hline
ExemptTot & Total tax exempted value of the building & 0.069 & 78.40\%\\
\hline
Radius\_Res\_Units\_Sold\_In\_Year & Residential units within 500 meters sold in past year & 0.056 & 79.60\%\\
\hline
\end{tabular}}
\end{table}

We observe that the regression model has a much higher dispersion of
feature importances compared to the classification model. The top
variable in the regression model, BuiltFAR, which is a measure of how
much of a building's floor to area ratio has been used (a proxy for
overall building size) contributes only 1.8\% of the reduction in the
error rate in the overall model. Conversely, in the classification
model, we see the top variable, ``Percent\_Neighbors\_Sold'' (a measure
of how many buildings within 500 meters were sold in the past year)
contributes 21.9\% of the total reduction in squared error.

Variable importance analysis of the regression model indicates that the
model favors variables which reflect building size (BuiltFAR, FacilFAR,
BldgDepth) as well as approximations for previous sale prices
(Last\_Sale\_Price and Last\_Sale\_Date). The classification model tends
to favor spatial lag features, such as how many buildings were sold in
the past year within 500 meters (Percent\_Neighbors\_Sold and
Radius\_Res\_Units\_Sold\_In\_Year) as well as characteristics of the
building function, for example, Percent\_Office, and Percent\_Storage.

\hypertarget{future-research-and-conclusions}{%
\section{Future Research and
Conclusions}\label{future-research-and-conclusions}}

\hypertarget{future-research}{%
\subsection{Future Research}\label{future-research}}

This research has shown that the addition of spatial lag features can
meaningfully increase the predictive accuracy of machine learning models
compared to traditional real estate valuation techniques. Several areas
regarding spatially-conscious machine learning models merit further
exploration, some of which we mention below.

First, it became apparent in the research that generalization was a
problem for some of the models, likely due to overfitting of the
training data. We corrected for this issue by employing more robust
algorithms; however, further work could be done to create variable
selection processes or hyperparameter tuning to prevent data
overfitting.

Additionally, the spatial lag features seemed to perform best for
certain boroughs and residential building types. We hypothesize that
using a 500-meter radius to build spatial lag features, a distance which
we arbitrarily chose, works best for this type of asset in these areas.
Fotheringham et al. (2015) used an adaptive bandwidth technique to
adjust the spatial lag radius based on cross-validation with much
success. The techniques presented in this paper could be expanded to use
cross-validation in a similar fashion to assign the optimal spatial lag
radius for each property.

Finally, this research aimed to predict real estate transactions 1 year
into the future. While this is a promising start, 1-year of lead time
may not be sufficient to respond to growing gentrification challenges.
Also, modeling at the annual level could be improved to quarterly or
monthly, given that the sales data contains date information down to the
day. To make a system practical for combating displacement, prediction
at a more granular level and further into the future would be helpful.

\hypertarget{conclusion}{%
\subsection{Conclusion}\label{conclusion}}

Societies and communities can benefit materially from gentrification,
however, the downside should not be overlooked. Displacement causes
economic exclusion, which over time contributes to rising income
inequality. Combating displacement allows communities to benefit from
gentrification without suffering the negative consequences. One way to
practically combat displacement is to predict gentrification, which this
paper attempts to do.

Spatial lags, typically seen in geographically weighted regression, were
employed successfully to enhance the predictive power of machine
learning models. The spatial lag models performed best for particular
building types and geographies; however, we feel confident that the
technique could be expanded to work equally as well for all buildings
with some additional research. Regarding algorithms, artificial neural
networks performed the best for predicting sale price, while gradient
boosting machines performed best for predicting sale occurrence.

While this research is not intended to serve as a full early-warning
system for gentrification and displacement, it is a step in that
direction. More research is needed to help address the challenges faced
by city planners and governments trying to help incumbent residents reap
the benefits of local investments. Income inequality is a complicated
and grave issue, but new tools and techniques to inform and prevent will
help ensure equality of opportunity for all.

\newpage

\hypertarget{references}{%
\section*{References}\label{references}}
\addcontentsline{toc}{section}{References}

\hypertarget{refs}{}
\leavevmode\hypertarget{ref-Dietzell2014}{}%
Alexander Dietzel, M., Braun, N., \& Schäfers, W. (2014).
Sentiment-based commercial real estate forecasting with google search
volume data. \emph{Journal of Property Investment \& Finance},
\emph{32}(6), 540--569.

\leavevmode\hypertarget{ref-Almanie2015}{}%
Almanie, T., Mirza, R., \& Lor, E. (2015). Crime prediction based on
crime types and using spatial and temporal criminal hotspots.
\emph{International Journal of Data Mining \& Knowledge Management
Process}, \emph{5}. \url{https://doi.org/10.5121/ijdkp.2015.5401}

\leavevmode\hypertarget{ref-antipov12}{}%
Antipov, E. A., \& Pokryshevskaya, E. B. (2012). Mass appraisal of
residential apartments: An application of random forest for valuation
and a cart-based approach for model diagnostics. \emph{Expert Systems
with Applications}.

\leavevmode\hypertarget{ref-Batty2013}{}%
Batty, M. (2013). The new science of cities. \emph{MIT Press}.

\leavevmode\hypertarget{ref-breiman2001random}{}%
Breiman, L. (2001). Random forests. \emph{Machine Learning},
\emph{45}(1), 5--32.

\leavevmode\hypertarget{ref-Quintos2013}{}%
Carmela Quintos PHD, M. (2015). Estimating latent effects in commercial
property models. \emph{Journal of Property Tax Assessment \&
Administration}, \emph{12}(2), 37.

\leavevmode\hypertarget{ref-Chapple2009}{}%
Chapple, K. (2009). Mapping susceptibility to gentrification: The early
warning toolkit. \emph{Berkeley, CA: Center for Community Innovation}.

\leavevmode\hypertarget{ref-Chapple2016}{}%
Chapple, K., \& Zuk, M. (2016). Forewarned: The use of neighborhood
early warning systems for gentrification and displacement.
\emph{Cityscape}, \emph{18}(3), 109--130.

\leavevmode\hypertarget{ref-Clay1979}{}%
Clay, P. L. (1979). \emph{Neighborhood renewal: Middle-class
resettlement and incumbent upgrading in american neighborhoods}. Free
Press.

\leavevmode\hypertarget{ref-Springer2017}{}%
d'Amato, M., \& Kauko, T. (2017). \emph{Advances in automated valuation
modeling}. Springer.

\leavevmode\hypertarget{ref-dimaggio2012spatial}{}%
DiMaggio, C. (2012). Spatial epidemiology notes: Applications and
vignettes in r. Columbia University press.

\leavevmode\hypertarget{ref-Dreier2004}{}%
Dreier, P., Mollenkopf, J. H., \& Swanstrom, T. (2004). \emph{Place
matters: Metropolitics for the twenty-first century}. University Press
of Kansas.

\leavevmode\hypertarget{ref-Eckert1990}{}%
Eckert, J. K. (1990). \emph{Property appraisal and assessment
administration}. International Association of Assessing Officers.

\leavevmode\hypertarget{ref-Fotheringham2015}{}%
Fotheringham, A. S., Crespo, R., \& Yao, J. (2015). Exploring, modelling
and predicting spatiotemporal variations in house prices. \emph{The
Annals of Regional Science}, \emph{54}(2), 417--436.

\leavevmode\hypertarget{ref-Friedman99stochasticgradient}{}%
Friedman, J. H. (1999). Stochastic gradient boosting.
\emph{Computational Statistics and Data Analysis}, \emph{38}, 367--378.

\leavevmode\hypertarget{ref-Fu2014}{}%
Fu, Y., Xiong, H., Ge, Y., Yao, Z., Zheng, Y., \& Zhou, Z.-H. (2014).
Exploiting geographic dependencies for real estate appraisal: A mutual
perspective of ranking and clustering. In \emph{Proceedings of the 20th
acm sigkdd international conference on knowledge discovery and data
mining} (pp. 1047--1056). ACM.

\leavevmode\hypertarget{ref-Geltner2017}{}%
Geltner, D., \& Van de Minne, A. (2017). Do different price points
exhibit different investment risk and return commercial real estate.

\leavevmode\hypertarget{ref-glass1964}{}%
Glass, R. (1964). \emph{Aspects of change}. London: MacGibbon \& Kee,
1964.

\leavevmode\hypertarget{ref-urban2016}{}%
Greene, S., Pendall, R., Scott, M., \& Lei, S. (2016). Open cities: From
economic exclusion to urban inclusion. \emph{Urban Institue Brief}.

\leavevmode\hypertarget{ref-Guan2014}{}%
Guan, J., Shi, D., Zurada, J., \& Levitan, A. (2014). Analyzing massive
data sets: An adaptive fuzzy neural approach for prediction, with a real
estate illustration. \emph{Journal of Organizational Computing and
Electronic Commerce}, \emph{24}(1), 94--112.

\leavevmode\hypertarget{ref-hastie01statisticallearning}{}%
Hastie, T., Tibshirani, R., \& Friedman, J. (2001). \emph{The elements
of statistical learning}. New York, NY, USA: Springer New York Inc.

\leavevmode\hypertarget{ref-Helbich2013}{}%
Helbich, M., Jochem, A., Mücke, W., \& Höfle, B. (2013). Boosting the
predictive accuracy of urban hedonic house price models through airborne
laser scanning. \emph{Computers, Environment and Urban Systems},
\emph{39}, 81--92.

\leavevmode\hypertarget{ref-hoffmann2004generalized}{}%
Hoffmann, J. P. (2004). \emph{Generalized linear models: An applied
approach}. Pearson College Division.

\leavevmode\hypertarget{ref-Johnson2007}{}%
Johnson, K., Benefield, J., \& Wiley, J. (2007). The probability of sale
for residential real estate. \emph{Journal of Housing Research},
\emph{16}(2), 131--142.

\leavevmode\hypertarget{ref-Silverherz1936}{}%
Joseph, D. S. (n.d.). The assessment of real property in the united
states. \emph{Special Report of the State Tax Commission, New York},
(10).

\leavevmode\hypertarget{ref-Kontrimasa2011}{}%
Kontrimas, V., \& Verikas, A. (2011). The mass appraisal of the real
estate by computational intelligence. \emph{Applied Soft Computing},
\emph{11}(1), 443--448.

\leavevmode\hypertarget{ref-Koschinsky2012}{}%
Koschinsky, J., Lozano-Gracia, N., \& Piras, G. (2012). The welfare
benefit of a home's location: An empirical comparison of spatial and
non-spatial model estimates. \emph{Journal of Geographical Systems},
\emph{14}(3), 319--356.

\leavevmode\hypertarget{ref-Lees2008}{}%
Lees, L., Slater, T., \& Wyly, E. (2013). \emph{Gentrification}.
Routledge.

\leavevmode\hypertarget{ref-Miller2015}{}%
Miller, J., Franklin, J., \& Aspinall, R. (2007). Incorporating spatial
dependence in predictive vegetation models. \emph{Ecological Modelling},
\emph{202}(3), 225--242.

\leavevmode\hypertarget{ref-Park2015}{}%
Park, B., \& Bae, J. K. (2015). Using machine learning algorithms for
housing price prediction: The case of fairfax county, virginia housing
data. \emph{Expert Systems with Applications}, \emph{42}(6), 2928--2934.

\leavevmode\hypertarget{ref-Pivo2011}{}%
Pivo, G., \& Fisher, J. D. (2011). The walkability premium in commercial
real estate investments. \emph{Real Estate Economics}, \emph{39}(2),
185--219.

\leavevmode\hypertarget{ref-Pollack2010}{}%
Pollack, S., Bluestone, B., \& Billingham, C. (2010). Maintaining
diversity in america's transit-rich neighborhoods: Tools for equitable
neighborhood change.

\leavevmode\hypertarget{ref-Rafiei2016}{}%
Rafiei, M. H., \& Adeli, H. (2015). A novel machine learning model for
estimation of sale prices of real estate units. \emph{Journal of
Construction Engineering and Management}, \emph{142}(2), 04015066.

\leavevmode\hypertarget{ref-Reardon2011}{}%
Reardon, S. F., \& Bischoff, K. (2011). Income inequality and income
segregation. \emph{American Journal of Sociology}, \emph{116}(4),
1092--1153.

\leavevmode\hypertarget{ref-Ritter2013}{}%
Ritter, N. (2013). Predicting recidivism risk: New tool in philadelphia
shows great promise. \emph{National Institute of Justice Journal},
\emph{271}.

\leavevmode\hypertarget{ref-Schernthanner2016}{}%
Schernthanner, H., Asche, H., Gonschorek, J., \& Scheele, L. (2016).
Spatial modeling and geovisualization of rental prices for real estate
portals. \emph{Computational Science and Its Applications}, \emph{9788}.

\leavevmode\hypertarget{ref-SCHMIDHUBER201585}{}%
Schmidhuber, J. (2015). Deep learning in neural networks: An overview.
\emph{Neural Networks}, \emph{61}, 85--117.
\url{https://doi.org/https://doi.org/10.1016/j.neunet.2014.09.003}

\leavevmode\hypertarget{ref-Smith1979}{}%
Smith, N. (1979). Toward a theory of gentrification a back to the city
movement by capital, not people. \emph{Journal of the American Planning
Association}, \emph{45}(4), 538--548.

\leavevmode\hypertarget{ref-turner_2008}{}%
Turner, H. (2008). Gnm: A package for generalized nonlinear models.
\emph{Department of Statistics University of Warwick, UK}. University of
Warwick, UK. Retrieved from
\url{http://statmath.wu.ac.at/research/friday/resources_WS0708_SS08/gnmTalk.pdf}

\leavevmode\hypertarget{ref-Turner2001}{}%
Turner, M. A. (2001). Leading indicators of gentrification in dc
neighborhoods: DC policy forum.

\leavevmode\hypertarget{ref-Watson2009}{}%
Watson, T. (2009). Inequality and the measurement of residential
segregation by income in american neighborhoods. \emph{Review of Income
and Wealth}, \emph{55}(3), 820--844.

\leavevmode\hypertarget{ref-Zuk2015}{}%
Zuk, M., Bierbaum, A. H., Chapple, K., Gorska, K., Loukaitou-Sideris,
A., Ong, P., \& Thomas, T. (2015). Gentrification, displacement and the
role of public investment: A literature review. In \emph{Federal reserve
bank of san francisco} (Vol. 79).

\end{document}